\RequirePackage{fix-cm}
\documentclass[smallcondensed]{svjour3}     
\smartqed  
\usepackage{natbib}
\usepackage{graphicx}
\usepackage{lineno,hyperref}
\modulolinenumbers[5]
\usepackage{amsmath}
\usepackage{balance}
\usepackage[byname]{smartref}
\usepackage{txfonts}
\usepackage{tocloft}
\usepackage{amsmath,amssymb,amsfonts} 
\usepackage{color}                    
\usepackage{comment}
\usepackage{mathtools}
\usepackage{commath}
\usepackage{graphics}
\usepackage[font={small,it}]{caption}
\usepackage{placeins}
\usepackage{tikz}
\usepackage{bchart}
\usepackage{enumitem}
\usepackage{adjustbox}
\usepackage{pstricks,pst-node,pst-tree}
\usepackage{pst-plot}
\usepackage{multirow}
\usepackage{epstopdf}
\AppendGraphicsExtensions{.tif}
\journalname{Artificial Intelligence Review}
\begin{document}

\title{Machine Learning Towards Intelligent Systems: Applications, Challenges, and Opportunities
}
\subtitle{}


\author{MohammadNoor Injadat        \and
        Abdallah Moubayed 			\and 
        Ali Bou Nassif				\and
        Abdallah Shami 
}


\institute{MohammadNoor Injadat, Abdallah Moubayed, Abdallah Shami \at
              Electrical \& Computer Engineering Dept.\\ 
              University of Western Ontario\\ 
              London, ON, Canada \\
              \email{minjadat@uwo.ca, amoubaye@uwo.ca, abdallah.shami@uwo.ca}           
            \and
            Ali Bou Nassif \at
            Computer Engineering Dept.\\ 
            University of Sharjah, Sharjah, UAE\\ 
            and\\
            Electrical \& Computer Engineering Dept.\\ 
            University of Western Ontario\\ 
            London, ON, Canada \\
            \email{anassif@sharjah.ac.ae}
}

\date{Received: date / Accepted: date}

\maketitle

\begin{abstract}
The emergence and continued reliance on the Internet and related technologies has resulted in the generation of large amounts of data that can be made available for analyses. However, humans do not possess the cognitive capabilities to understand such large amounts of data. Machine learning (ML) provides a mechanism for humans to process large amounts of data, gain insights about the behavior of the data, and make more informed decision based on the resulting analysis. ML has applications in various fields. This review focuses on some of the fields and applications such as education, healthcare, network security, banking and finance, and social media. Within these fields, there are multiple unique challenges that exist. However, ML can provide solutions to these challenges, as well as create further research opportunities. Accordingly, this work surveys some of the challenges facing the aforementioned fields and presents some of the previous literature works that tackled them. Moreover, it suggests several research opportunities that benefit from the use of ML to address these challenges.
\keywords{Machine learning \and Data Analytics \and Application fields \and Research
	Opportunities}
\end{abstract}

\section{Introduction} \label{intro}
\indent The rapid growth of the Internet and related technologies has provided individuals, organizations, and society with the opportunity to collect large amounts of data \citep{s1}. However, these large amounts of data often lead to information overload. Information overload occurs when the amount of input (e.g. data) that a human is trying to process exceeds their cognitive capacities \citep{s2}. Information overload can lead to humans ignoring, overlooking, or misinterpreting crucial information \citep{s3}.\\ 
\indent Humans do not have the cognitive capacity to process large amounts of data. Therefore, the discipline of data science has emerged. Data science combines the classic disciplines of statistics, data mining, databases, and distributed systems in order to extract information from large sets of data \citep{s1}. One approach of data analysis that data scientists can implement is machine learning (ML) \citep{Moubayed_thesis}. ML allows computers to learn without being explicitly programmed. Once the computer learns patterns from a training set of data, it can apply what it has learned to find these patterns in similar data \citep{s4}. Furthermore, ML allows computer systems to adapt and learn from their experience \citep{s5,s6}.\\
\indent ML algorithms have a lot of applications. Examples are house pricing prediction, spam filtering, education, structuring of data in healthcare systems, drug response prediction, diabetes research, network security, banking and finance, and social media. This work aims to provide a brief literature review of the challenges facing different fields such as education, healthcare, network security, banking and finance, and social media. Moreover, it presents several research opportunities on the role and potential of using ML to address these challenges. Hence, the contributions of this paper are summarized as follows:

\begin{itemize}
	\item Describing briefly the different challenges facing a variety of modern fields including education, healthcare, network security, banking and finance, and social media.
	\item Presenting some of the previous literature works that addressed these challenges and their shortcomings.
	\item Discussing the role and potential of ML in addressing these challenges and presents potential frameworks for its deployment.
\end{itemize}
\indent The remainder of this paper is organized as follows: Section \ref{ML_trends} presents some of the recent trends concerning the development and deployment of ML algorithms. Section \ref{educ} discusses the education field. Section \ref{healthcare} focuses on the healthcare system and field. Section \ref{network_sec} presents the challenges in network security and the potential role of ML in addressing these challenges. Section \ref{buss_finance} sheds light on the banking and finance sector. Section \ref{Social_Media} focuses on the area of social media. Finally, Section \ref{conclusion} concludes the paper.
\section{Recent Trends in Machine Learning}\label{ML_trends}
\begin{figure}[!t]
	\centering
	\includegraphics[scale=0.4]{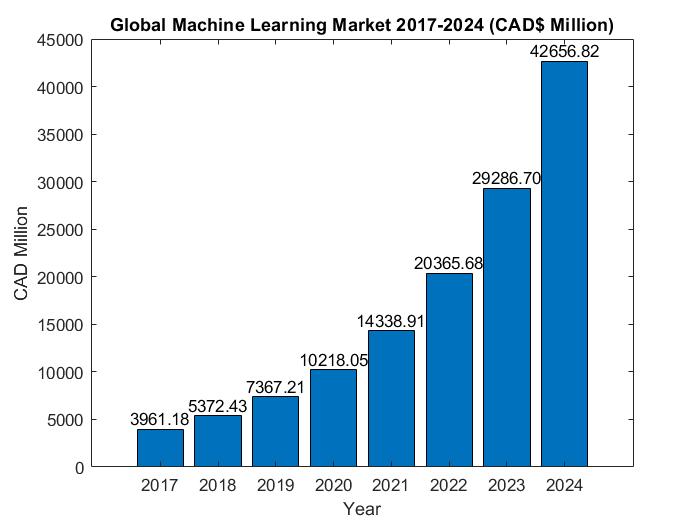}
	\caption{\label{ML_forecast} Global ML Forecast \cite{ML_adoption_statistic}}
\end{figure}

\indent\indent ML has become an extremely popular topic within development organizations that are looking to adopt a data-driven approach to improve their business by gaining useful information from the data they collect. With ML models, organizations can continually predict changes in their business and make decisions accordingly. ML uses algorithms that iteratively learn from data to improve, describe data, and predict outcomes. Once an ML model has been trained, it can predict new data that is given as input. The output given by the model on the new data will depend on the data used to train the model.\\
\indent The emerging growth of ML adoption in various fields is emphasized by the amount of financial resources being allocated to deploy ML models. As illustrated in Figure \ref{ML_forecast}, the global ML market is expected to reach close to \$42.5 billion CAD by the year 2024 \citep{ML_adoption_statistic}. Furthermore, as per McKinsey \& Company's 
``Notes from the AI Frontier, Tackling Europe’s Gap in Digital and AI'' discussion paper, the ML market could boost economic activity growth throughout the EU by as much as 20\% by the year 2030 \citep{ML_adoption_statistic2}. Moreover, the World Economic Forum predicted that a net of 58 million jobs will be created in the coming years due to ML technologies \citep{ML_adoption_statistic3}. This highlights the importance and positive potential impact that ML will have on the global economic market.\\
\indent Within the area of ML, one promising paradigm to adopt is federated learning (FL). FL is a ML paradigm in which a high-quality centralized model is trained using data that is distributed over a large number of locations. The term was first coined by Google in 2016 in which they proposed a mechanism in which data at each location is used to independently compute an update of the current ML model. This update is then communicated back to a central service that aggregates these updates to compute a new global model that is distributed back to the different locations \citep{FL_google}. Accordingly, this paradigm adopts the ``bringing the code to the data'' philosophy rather than ``bringing the data to the code''. As such, the FL paradigm addresses concerns regarding the data privacy, ownership, and locality \citep{FL_google2} \citep{FL_yang}. Given the distributed nature of water leak monitoring systems with sensors collecting data at various geographical locations, FL promises to be a viable solution for extracting meaningful information from the collected data while still maintaining its privacy and locality.\\
\indent The continued projected market growth of ML technologies and the privacy-preserving characteristic of FL (due to its distributed learning nature) has resulted in increased demand for FL with new technologies and frameworks currently being developed. For example, Google has recently released a TensorFlow-based FL framework named TensorFlow Federated (TFF). TFF enables developers to deploy an AI system and train it across data from multiple sources, all while keeping each of those sources separate and local \citep{FL}. Other FL-based frameworks include Federated AI Technology Enabler \citep{FL_AI}, PySyft \citep{pysyft}, Leaf \citep{leaf}, PaddleFL \citep{Paddle_FL}, and Clara Training Framework \citep{NVIDIA_clara}. In addition to the popular ML algorithms previous proposed in the literature such as neural networks and support vector machines, this illustrates the recent and continued development and deployment efforts of ML algorithms and paradigms.
\section{Education}\label{educ}
\indent The first area considered is that of the education sector. There are three main ways that education can be delivered: onsite, online, and blended learning. Onsite education, or traditional education, refers to educational content delivery within a traditional classroom setting \citep{s11}. This setting requires that the educator and the students are in the same room at the same time. This allows the educator to deliver his/her lecture to the attending class. As such, traditional classrooms provide face-to-face interaction between the educator and the students \citep{s13}.\\
\indent On the other hand, online education, one category of e-learning systems, refers to education that is provided over the Internet. E-learning provides students with the opportunity to access educational curriculum outside of a traditional classroom at any time from any geographical location. However, there is no face-to-face interaction with the educator as all the content is delivered remotely \citep{s11}. \\ 
\indent Last but not least, blended or hybrid learning is a combination of onsite and online education. For the education delivery system to be considered blended, up to 30\% of the course requirements must be conducted face-to-face in a traditional classroom setting, while the remaining percentage of the course requirements can be completed online. Blended learning offers students the opportunity to have face-to-face interactions with the educator and other students while also providing them with the opportunity to assess course materials at any time from any location \citep{s11}.\\ 
\indent However, the educational sector faces a variety of challenges, some of which are pedagogical and others being technical. This section identifies some of the challenges in the education sector. Moreover, it presents some of the previous literature works that tried to address each of these challenges. Furthermore, it discusses the role of ML in addressing them. challenges. More specifically, this section will discuss how ML can be used to grade essays, predict and prevent students from dropping-out, improve intelligent tutoring systems, recommend online courses, and provide personalized learning. 
\subsection{Essay Grading}
\subsubsection{Challenge Description:}
\indent Essays provide a tool for assessing students’ critical thinking, analysis, and communication skills. However, it is time consuming for educators to grade essays. Furthermore, when humans grade essays there is a great level of subjectivity which can lead to two different graders scoring an essay very differently \citep{s15}. Within this context, using ML algorithms to grade essays can reduce the workload of educators and provide more objectivity during the grading process. A common approach to creating an essay grading algorithm is to first collect a large pool of essays which have characteristics that are computationally measurable (e.g., sentence length, word frequency distributions, grammar, and spelling), and have been scored by humans \citep{s16}. This allows the algorithm to first learn the characteristics which are important for grading an essay. Then, when the algorithm is used to score essays, the algorithm’s scores can be compared to those of the human graders in order to determine if the algorithm has properly learned the grading characteristics. It is worth mentioning that this challenge is not only from the technical aspect, but also from the social aspect in terms of accepting the output of the automated ML-based models. However, this work only focuses on the technical aspect of this challenge rather than the social aspect of it.
\subsubsection{Previous Works:} 
\indent \citet{s15} built an automated essay scoring system using essays from kaggle.com. The authors selected roughly 13,000 essays from a pool of essays that were submitted to a competition by the William and Flora Hewlett Foundation. The essays were written by students from Grade 7 to Grade 10 and were approximately 150 to 550 words long. The selected essays were divided into eight sets, with each of the sets having unique grading characteristics. Eight different sets were selected to ensure that the automated grader was trained across different types of essays. Furthermore, each essay has one or more human scores. In the latter case, the essay also had a final resolved score which considered all human scores. After that, the authors selected the eight sets of training essays, they extracted several features from them (e.g., total word count per essay, sentence count, number of long words, part of speech counts, etc.) . These features were selected because they are characteristics that a human grader would commonly look for when grading an essay. The authors then used a linear regression model to allow their algorithm to learn parameters for grading based on the selected features. After the algorithm learned the parameters for scoring the eight different essay types, the algorithm was used to score a distinct set of test essays. These scores were compared against human graded scores to arrive at an error metric (Quadratic Weighted Kappa). The average kappa score of the authors’ algorithm across the eight essay types was 0.73. Essay set 8 had the lowest kappa at 0.68 and essay set 1 had the highest kappa at 0.80. Despite the fact that the proposed framework achieved good performance as seen with the high kappa values, this work has a limited contribution as it only considered one ML algorithm. \\
\indent \citet{s16} also used ML techniques to develop an automated essay assessment system . The authors selected essays from a pool of essays that were submitted to a competition by The Hewlett Foundation to kaggle.com. All of the essays had been graded by humans. Similarly, the authors further segregated these essays into eight unique sets. This was followed by using Bayesian Linear Regression as their algorithm. The Bayesian essay scoring system used features like specific words, specific phrases, order in which certain noun-verb pair appears, and the order of the concepts explained to score the essays. In the end, the authors tested their algorithm on eighty essays divided into two groups of forty, which had also been scored by humans. The authors’ algorithm was over 80\% accurate at scoring the essays. Yet, this work also only considered one ML technique which limits its contribution.\\
\indent \citet{s14} also discussed using ML to assess essays. More specifically, the author investigated the potential of ML algorithms to automate the analysis of reflective essays. To that end, the author explored eight different categories that are often used as metrics to evaluate the quality of a reflective passage, namely reflection, experience, feeling, belief, difficulty, perspective, learning, and intention. To that end, the author collected data from 76 students containing a total of 5080 sentences. Then, the authors created a training dataset using a random sample consisting of 80\% of the sentences and tested the performance of four different ML classification algorithms including support vector machines (SVM), neural networks, random forests (RF), and Naive Bayes on the remaining 20\% of the sentences. Experiments showed that the accuracy of the different ML models ranged between 80\%-96\% for the different reflective essay metrics.\\
\indent \citet{AEG_DL} explored the use of deep learning models to automate the essay grading process. To that end, the authors used the ASAP AEG dataset that described differed essay sets with multiple essay traits such as Content, Organization, Word Choice, Sentence Fluency, and Conventions \citep{AEG_dataset}. Moreover, the authors proposed the use of a feature engineering system, a string kernel, and an attention-based neural network as part of their automated essay grading framework. Additionally, the Cohen's Kappa metric was used to evaluate the performance of the proposed framework as it considers the random correct classification of data samples. Results showed that the attention-based neural network algorithm outperformed other works from the literature across multiple prompts (with each prompt having a different set of essay traits) with a Kappa value between 0.586 and 0.820. However, the work only considered deep neural networks without considering other potential classification algorithms that may be more computationally efficient and have similar performance.
\subsubsection{Research Opportunities:}
\indent As can be seen from this review of the literature, essays are an important tool for assessing students’ comprehension and expression. However, grading essays is a time consuming task. Furthermore, essay grading is prone to subjectivity, which can lead to the same essay being scored differently by two graders. Hence, essay grading presents the challenges of time consumption and human subjectivity.\\ 
\indent ML offers a potential solution to address these issues. Firstly, ML algorithms can be used so that graders no longer need to spend time on grading. Secondly, such algorithms can be used to provide objective scores of essays. Although the previous subsection presented research that has shown how ML can be used to address the challenges of essay grading, there are still opportunities for further research.\\ 
\indent One potential opportunity is exploring and evaluating different ML models (e.g. logistic model trees or deep neural networks). This is mainly due to the fact that most of the previous work only used one algorithm for essay grading. Therefore, it is important to explore and compare the performance of other ML models to obtain a more robust essay grading framework, especially given the effectiveness of other models such as deep neural networks in natural language processing problems. Another potential opportunity is studying the impact of more advanced Natural Language Processing features (e.g. N-grams, k-nearest neighbors (k-NN) in bag of words), selecting features that are grammar and usage specific, and exploring other polynomial basis functions like neural networks (NN) as part of the essay grading framework. \\
\indent Such frameworks can be applied to any assessment task that contains an essay component. This includes exams and tests that contain essay sections. The application of essay grading algorithms to exam and test essays could increase the consistency of scoring while reducing the grader bias. Furthermore, there is the possibility to use essay grading algorithms as components of interactive knowledge and writing tutorial systems.
\subsection{Dropout Prevention}
\subsubsection{Challenge Description:}
\indent Student dropout is another challenge that is prevailing in the education sector. The term dropout refers to the case when a student leaves/quits a course before completing it. Recent studies showed that students were more likely to dropout of online courses than traditional classes \citep{s17}. High dropout rates can effect the future of colleges and universities. This is because different stakeholders within the education field including policymakers, funding bodies, and educators consider dropout rates to be an objective outcome-based measure of the educational institutions' quality \citep{s18}.\\ 
\indent While the higher dropout rate of students in online classes is a known issue, there are many possible reasons for students to dropout. In turn, this makes predicting dropout challenging. From the student side, possible reasons for online course dropout include higher than expected workload, inability to manage academic responsibilities in a self-driven learning environment, unfamiliarity with the online educational delivery system, less student–teacher interaction, family and social obligations, and motivation level \citep{s20}. However, students may also dropout of online courses due to the course being poorly designed and delivered, which can occur when the professor who created and taught the online course is unfamiliar with technology and/or is provided with no training by their institution on how to teach in an online environment \citep{s20}.\\ 
\indent As can be seen, there are many possible reasons students may dropout, which can make predicting which students will dropout a complicated task. Even though this is a complicated task, it is important to identify students at risk of dropping out so that professors can address the needs of these students and take the appropriate actions to reduce their probability of dropping out \citep{s18}. One way to make the task of identifying at risk students easier is to use ML algorithms. Again, this challenge does not only entail the technical aspect, but it also has a pedagogical aspect in terms of the context used for data collection purposes. However, as mentioned earlier, this work only focuses on the technical aspect with which ML can play a role without tackling the pedagogical context of the data collection.
\subsubsection{Previous Works:} 
\indent \citet{s17} used ML techniques in order to predict dropout in e-learning courses. More specifically, the authors proposed the use of logit leaf model (LLM), a decision tree (DT)-based classification model, to accurately predict student dropout in subscription-based e-learning environments. LLM was chosen due to its capability to balance between comprehensibility and predictive performance. To that end, the authors compared the performance of the proposed LLM model to that of eight other ML classification algorithms on a real-life dataset containing more than 10,000 students of a global subscription-based e-learning provider. Results showed that the proposed LLM model was one of the top performing student dropout prediction models with a high area under the curve (AUC) value above 0.8, highlighting its effectiveness in achieving its target task. \\
\indent Similarly, \citet{s21} proposed the use of an ML classification model to predict student dropout in high schools. In particular, they proposed the use of RF algorithm for this task due to its high prediction accuracy in multiple scenarios and applications. To that end, the authors used the National Education Information System (NEIS) data collected in Korea in 2014 and evaluated the performance of the RF model using multiple metrics such as accuracy, sensitivity, specificity, and AUC. Experiments showed that the proposed RF achieved high accuracy (close to 95\%) and high AUC value (close to 0.97), highlighting the effectiveness of this model. However, one shortcoming is that the work only considered one classification algorithm without comparing it to other potential ML algorithms.
\subsubsection{Research Opportunities:}
\indent Although ML has been proposed to predict dropout in e-learning courses, there are still research opportunities within this area. One potential opportunity is studying the performance of different ML dropout prediction frameworks and models in other course delivery settings such as blended learning, distance and classical education. This would highlight the generality of the dropout prediction framework. Another opportunity worth exploring is investigating the impact of different student attributes to create their dropout prediction method. This is essential as it can result in more accurate models. A third opportunity to consider is comparing the performance of different base and ensemble learning methods to achieve more accurate and robust prediction models and studying their impact on retention strategies through correlation and association rules mining. 
\subsection{Intelligent Tutoring}
\subsubsection{Challenge Description:}
\indent The third challenge facing modern education systems is that of providing and improving intelligent tutoring systems \citep{s22}. For example, \citet{s23} proposed an intelligent tutoring system to teach English and French languages using ML-based models. However, in such a scenario, there are multiple challenges that arise include how to detect spelling mistakes, verb tense mistakes, and auxiliary verb mistakes. As such, developing effective intelligent tutoring systems can be a challenging task given the significant impact of multiple factors that need to be considered. 
\subsubsection{Previous Works:} 
\indent \citet{s22} combined the concepts of ML and gamification with cloud technologies in a unified framework to improve intelligent tutoring systems. More specifically, the authors proposed the use of different ML models such as optical character recognition, sentiment analysis, and speech recognition to create a \textit{virtual study buddy}. The goal of this system is to help students develop better study strategies by interacting with a digital study partner. However, one limitation of this work was the fact that the authors did not test their proposed framework to explore its effectiveness.\\
\indent On the other hand, \citet{s24} proposed the use of sentiment analysis to improve the performance of an affective intelligent tutoring system by better gauging the opinions of students about the course contents. To that end, the authors used a collection of texts containing more than 68,000 Twitter messages written in Spanish and transformed them into numerical feature vectors along with their associated sentiment. Then, the authors used Naive Bayes classifier (chosen due to its simplicity) to predict the sentiment of future texts/Twitter messages. Again, the authors used multiple metrics such as accuracy, precision, recall, and f1-score to evaluate the performance of the proposed module. Experimental results showed that their proposed module achieved high accuracy (above 80\%) coupled with high f1-scores. However, one shortcoming of this work is the fact that they did not compare the performance of their proposed module to other potential classifiers.
\subsubsection{Research Opportunities:}
\indent Despite the fact that ML has been used to improve intelligent tutors, more research opportunities still exist. One such opportunity is considering more features (for example phonemic and history-based features) and investigating their impact on the performance of the developed model. Another potential opportunity to consider is comparing the performance of other classifiers such as NN and SVM. This comparison can help determine whether the classifiers previously proposed in the literature are biased. Moreover, such comparisons will lead to having a more adaptive and robust intelligent tutors. 

\subsection{Course Recommendation}
\subsubsection{Challenge Description:}
\indent Massively Open Online Courses (MOOCs) such us Coursera, Udacity, EdX, and MOOC.org are a form of online distance education/e-learning. As such, MOOCs provide online courses that can be accessed by a student at any time from any geographical location \citep{s11}. MOOCs are open to anyone that is interested in enrolling and are often free or low-cost. However, the courses do not provide course credit and are not applied towards a degree. Instead, MOOCs tend to be used by people who want to learn new skills, be it to advance their career or for fun. MOOCs provide open access to a plethora of courses from various top-rated universities and institutions. For example, the website mooc.org provides a course titled “Data Science: R Basics” from Harvard University, and a course titled “Introduction to Data Analysis using Excel” from Microsoft \citep{s25}.\\
\indent Since MOOCs are open access, hundreds of thousands of students can be enrolled in each course, with MOOCs platforms offering thousands of different courses. This means that MOOC platforms are privy to mass amounts of data. This data can then be used to improve the MOOC system. For example, having thousands of different courses available can be overwhelming for students. Therefore, if a student is looking to improve a specific skill, it would be beneficial for the MOOC system to recommend which courses are needed to acquire those skills \citep{s26}.
\subsubsection{Previous Works:} 
\indent Several previous literature works focused on the problem of course recommendation for students. One such example is the work by \citep{s27}. The authors used prior student data and a combination of ML algorithms to recommend courses to students in an e-learning system. The authors combined Simple K-means (a clustering technique) and Apriori (an association rule algorithm) to investigate prior students’ data from Moodle.org in order to determine which courses to recommend to new students. The authors found that the results of their combination approach matched real world student course selection patterns. However, one limitation of this work is that it only considered one unsupervised clustering algorithm.\\
\indent \citet{s27a} also proposed the use of ML algorithms as part of a course recommendation system for online learning environments. More specifically, the authors first used K-means algorithm to group students based on their performance in previous courses. This was followed by applying collaborative filtering to recommend new suitable courses. The results showed that the proposed model achieved a low root mean squared error and mean absolute error. Furthermore, the model also achieved high precision and recall values, indicating that it can return correct results and preserve the majority of true positives.\\
\indent Similarly, \citet{s27b} proposed the use of a distributed association rule algorithm as part of their course recommendation system. The authors used a combination of Hadoop and Spark platforms to implement the proposed framework so that it is suitable for MOOC environments. The experimental results on three different datasets illustrated the effectiveness of the proposed framework by having a high confidence value (close to 0.5) for multiple association rules.
\subsubsection{Research Opportunities:}
\indent ML algorithms can be further applied to the large amounts of data that MOOC platforms possess in order to determine which courses would be best for a student who is interested in improving a specific skill set. One potential research opportunity for students' course recommendation is evaluating the courses that other students have taken that are related to the skill that the student is interested in. Using that information can help build an effective course recommender. Another opportunity is consider multiple supervised classification algorithms. This is particularly important given the substantial impact that the classification process has on the overall performance of the recommender. Therefore, it is worth exploring the performance of different classification algorithms to study their impact on the effectiveness of the recommendation process. 
\subsection{Personalized Learning}
\subsubsection{Challenge Description:}
\indent Personalized learning is based on the individual students and how they learn. Each individual learns differently and has a unique learner profile. This profile is based on the individual’s learning style \citep{s28,s29,s30}, which consists of specific behaviors and attitudes \citep{s31}. Personalizing each learner’s education can lead to better learning. One way to personalize education is by using recommender systems that provide useful suggestions for users (books, movies, products, etc.) based on their preferences and their similarity to other students \citep{s32}.
\subsubsection{Previous Works:} 
\indent \citet{s32} tested the ability of LearnFitII to act as a recommender system. LearnFitII is an adaptive learning system that automatically adapts to the dynamic preferences of learners. By mining the server logs of students, LearnFitII was able to recognize the different learning styles and habits of students. Then, using the Felder-Silverman model of learning styles, LearnFitII proposed personalized learning scenarios. The Felder-Silverman model of learning styles consists of four learning dimensions (1. Information Processing, 2. Information Perception, 3. Information Reception, and 4. Information Understanding) \citep{s33}. These dimensions can be accessed via the Index Learning Style Questionnaire (ILSQ) which consists of 44 questions.\\ 
\indent After proposing personalized learning scenarios, LearnFitII analyzed the habits and the preferences of learners by mining information about the learners’ actions and interactions. After the mining of this information, the learning scenarios were revisited and updated using a hybrid recommender system which combined k-NN and association rule mining algorithms. The authors found that when LearnFitII was tested in real environments that learning quality increased and so did the learners’ satisfaction with the learning process \citep{s32}.\\ 
\indent Another way that personalized learning can be beneficial is in helping students select the learning-pathway that is appropriate for them. \citet{s34} investigated how student learning-pathways can be improved with ML. The term learning-pathway can be understood as the path of academic courses that is appropriate for a student to achieve a degree. Ideally, one’s learning-pathway is in their field of interest. Typically, students spend some time taking various courses in order to discover which topics they are interested in. However, this process of taking various courses can lead to a mismatch between a student’s current and preferred learning pathway. When mismatches occur, the student may experience academic difficulties (e.g. weak performance, high absentee rate). These mismatches may lead students to lower their level of education or dropout of university altogether. In order to improve students’ levels of achievement it would be beneficial to help them determine their desired learning-pathway sooner. In order to achieve this goal sooner, the authors first collected questionnaire data. Then they sent a questionnaire to 900 students from the Faculty of Computers and Information Technology at Tabuk University in Saudi Arabia with 450 students returning the questionnaire. The questionnaire addressed four topics: basic information, personal information, academic information, and learning pathway information. After collecting this data, the authors applied a DT algorithm to the data. Then, induction rules were deduced from the tree paths in order to provide learning-pathway recommendations. In order to validate their results, the authors divided the questionnaires into two groups, a developing group (70\%) and a test group (30\%). Using these two groups of data, the authors found that their algorithm could accurately provide learning-pathway recommendations \citep{s34}.\\
\indent \citet{sa,sb} studied the problem of student engagement level identification in an e-learning environment. This was done using K-means algorithm. In addition to that, the authors extracted a set of rules relating student engagement to academic performance. To that end, the authors used the Apriori association rules algorithm. Experimental results showed a positive relationship between students' engagement level and their academic performance.\\
\indent \citet{injadat_ch4,injadat_ch5} proposed the use of optimized ML ensemble classification models to predict student performance during the course delivery time at two stages. The authors explored the use of various based learners such as SVM, K-NN, NB, RF, and neural networks to form the ensembles and tested them on two different datasets. Results showed that their proposed ensemble models achieved high accuracy for the target class in both the binary and multi-class cases despite the small number of instances available.
\begin{table}[!htbp]
	\centering
	\caption{\label{ML_educ_table} Challenges, Previous Works, and Research Opportunities within Education Sector}
	\scalebox{0.9}{
	\begin{tabular}{|p{2.5cm}|p{4cm}|p{4cm}|}
		\hline
		\textbf{Challenge} & \textbf{Previous Work} & \textbf{Research Opportunity}  \\ \hline
		\multirow{4}{2.5cm}{Essay Grading} & Regular linear regression is used for essay grading \citep{s15} & - Explore different ML models such as LR, DT, and DNN\\ \cline{2-2}
		& Bayesian linear regression is used for essay grading \citep{s16} & - Study the impact of additional Language and usage-specific features as well as other polynomial basis functions.\\ \cline{2-2}
		& SVM, RF, neural networks, and naive bayes are used to evaluate reflective essays \citep{s14} & - Compare the performance of the different models on various tasks to get a more accurate and robust essay grading framework. \\ \cline{2-2}
		& Deep neural networks were used to classify essays based on five different potential traits \citep{AEG_DL}&\\ \hline
		\multirow{3}{2.5cm}{Dropout Prevention}& Studied the performance LLM to predict dropout \citep{s17}& - Study the performance of different models (base learners and ensemble learner models) in different course delivery settings. \\ \cline{2-2}
		&Studied the performance of RF classification model for dropout prediction\citep{s21}& - Investigate the impact of different student attributes on the dropout prediction frameworks.\\ \hline
		\multirow{2}{2.5cm}{Intelligent Tutors}& Sentiment analysis was proposed as part of a virtual study buddy framework \citep{s22}& - Consider more features such as phonemic and history-based features as well as investigate their impact on the performance of the developed model. \\ \cline{2-2}
		&NB was used to predict the sentiment of text messages to be used to improve an affective intelligent tutoring system \citep{s24}& - Study the performance of other classifiers such as NN and SVM to determine whether the classifiers previously proposed in the literature are biased.\\ \hline
		\multirow{3}{2.5cm}{Course Recommendation}& Combined k-means and apriori algorithm to recommend courses \citep{s27}& - Evaluate the courses that other students have taken that are related to the skill the student is interested in using multiple metrics. \\ \cline{2-2}
		&Used a combination of K-means algorithm and collaborative filtering for course recommendation \citep{s27a}& - Consider multiple supervised classification algorithms to study their impact on the effectiveness of recommendation process.\\ \cline{2-2}
		&Used an improved version of Apriori association rules algorithm for course recommendation \citep{s27b}&\\ \hline
		\multirow{5}{2.5cm}{Personalized Learning}&Combined a DT algorithm and induction rules algorithm to provide learning-pathway recommendations \citep{s34}& - Study the performance of different classification algorithms to predict student performance during the course delivery. \\ \cline{2-2} 
		& Mined server logs to determine student learning style \citep{s32}& - Consider more complex recommendation approaches by including other factors.\\ \cline{2-2} 
		&Used K-means and apriori algorithms to identify student engagement and their relation with academic performance \citep{sa,sb} & \\ \cline{2-2} 
		&Used multiple optimized ensemble classification algorithms to predict student performance during course delivery \citep{injadat_ch4,injadat_ch5}& \\ \hline
	\end{tabular}}
\end{table} 
\subsubsection{Research Opportunities:}
\begin{figure}[!t]
	\centering
	\includegraphics[scale=0.3]{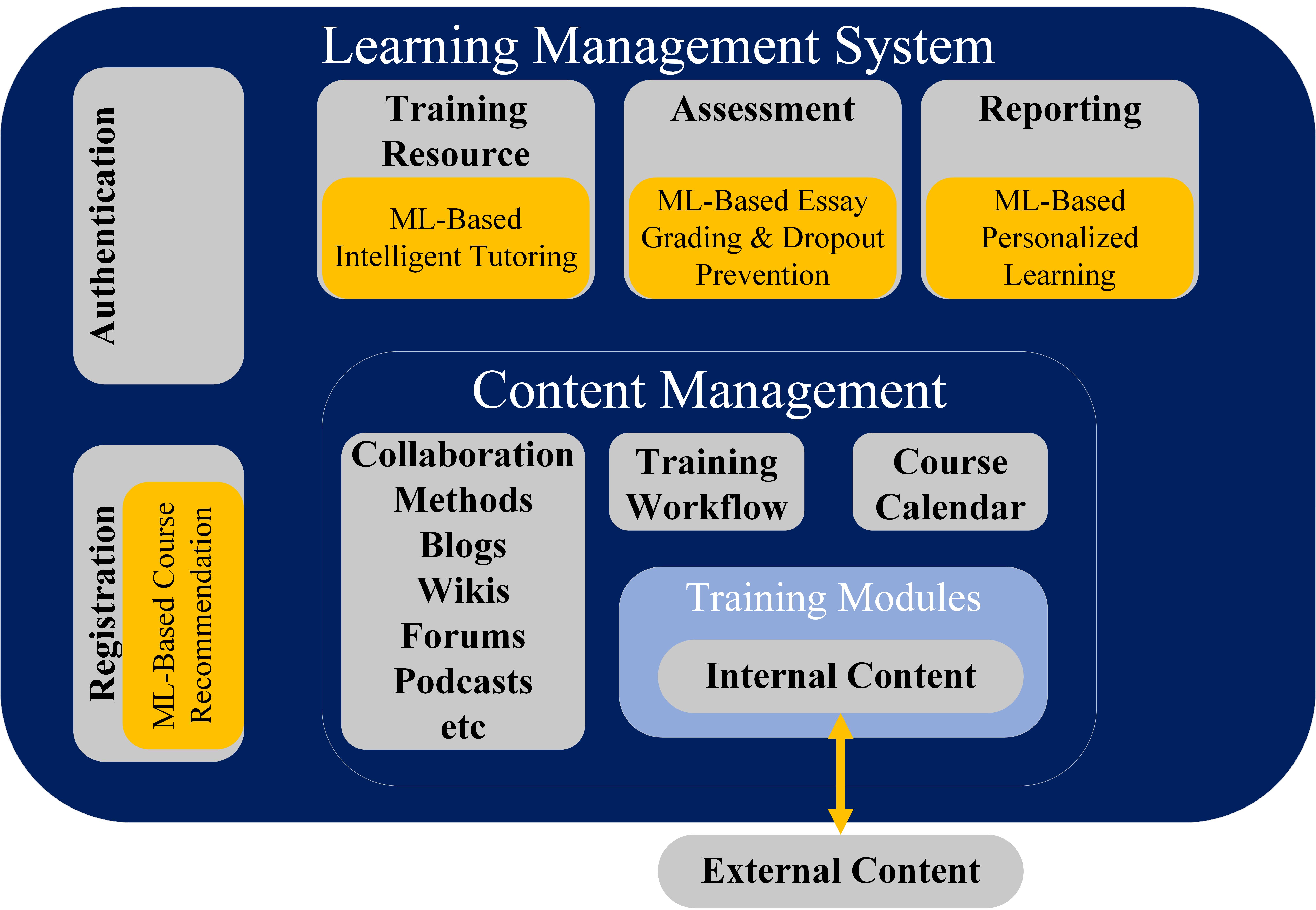}
	\caption{Potential Deployment of ML in Learning Management System (LMS)}
	\label{ML_in_education} 
\end{figure}
\indent There are still further research opportunities to use ML to provide personalized learning. One opportunity is to consider more complex recommendation approaches by including other factors such as learner motivation and knowledge level as well as additional personality traits. Another opportunity is to study the performance of different classification algorithms to predict student performance during the course delivery. This can help identify student who may need help and provide them with a personalized plan to improve their predicted performance.\\
\indent Table \ref{ML_educ_table} summarizes the challenges within the education sector, lists some of the previous works, and presents the different research opportunities. Furthermore, Figure \ref{ML_in_education} illustrates the potential deployment framework of the ML modules within the Learning Management System (LMS).
\section{Healthcare}\label{healthcare}
\indent Another area where ML has shown promise is in the field of healthcare. Many modern medical organizations use electronic health records (EHRs) \citep{s3}, EHRs consist of heterogeneous data elements. This includes information such as the patient demographic, diagnoses, laboratory test results, previous prescriptions, and clinical notes \citep{s36}. Patient data can also include imaging, sensor and text data \citep{s37}. Furthermore, this data often comes in various formats, including structured, semi-structured and weakly structured data \citep{s38}. Originally it was thought that having access to more information about individual patients would lead to more informed medical decisions. However, often times health professionals are overwhelmed by the amount of information that is now available to them \citep{s3}. Hence, a challenge with big data in healthcare is making the data easily interpretable for medical professionals. ML offers a solution to this problem because it can be used to identify relevant patterns in complex data. In this section, how ML algorithms can be used in various applications such as predicting individual patient’s responses to cancer drugs, diabetes research, retinopathy detection, and cancer detection is discussed. Note that despite the various applications in which ML can be applied within the field of healthcare, this section discusses some of the most prominent applications.
\subsection{Drug Response Prediction}
\subsubsection{Challenge Description:}
\indent One way that ML can be applied to medical data is to predict an individual patient’s response to a drug or drugs \citep{s39}. For example, ML can be used to predict the responses of individual cancer patient to therapeutic drugs \citep{s40}. When working with cancer patients, it is possible to use precision cancer medicine. Precision cancer medicine aims to accurately predict the optimal drug therapies for a patient based upon the personalized molecular profiles of their tumors \citep{s41}. In order to provide precision cancer medicine, it is necessary to search for significant correlations between patient tumor profiles and the output predictions of optimal drug responses in cancer-relevant datasets. Once these correlations are found in previously established datasets, they can be used to predict an individual patient’s response to various series of therapeutic drugs \citep{s39}. As mentioned before, ML offers a solution to this problem, because it can be used to identify relevant patterns in complex data. 
\subsubsection{Previous Works:} 
\indent \citet{s40} applied their open-source SVM-based algorithm to the gene-expression profiles of 175 individual cancer patient’s tumors. The algorithm was able to predict the responses of these 175 individuals to a variety of standard-of-care chemotherapeutic drugs with $>$80\% accuracy.\\
\indent \citet{s42} also used ML to predict tumor cell line response to drug pairs. The authors used a computational deep learning model to predict cell line response to a subset of drug pairs in the National Cancer Institute-ALMANAC database. When the authors ranked the drug pairs for each cell line based on the model’s predicted combination effect, they were able to determine 80\% of the top drug pairs.\\ 
\indent \citet{s43} also used deep neural networks (DNN) and the genomic profiles of cancer tumors in order to predict the tumors’ responses to therapeutic drugs. The authors created DeepDR, a deep neural network model, then trained it to learn the genetic background of tumors based on data from The Cancer Genome Atlas (TCGA). DeepDR was also trained on pharmacogenomics data from human cancer cell lines provided by the Genomics of Drug Sensitivity in Cancer (GDSC) Project. After training on these data sets, DeepDR was applied to TCGA data again in order to predict the drug response of tumors. The authors’ work provides insights into the ability of a deep neural network model to translate pharmacogenomics features identified from in vitro drug screening to predict the response of tumors.\\
\indent \citet{s44} used ML and genetic data to predict patients’ responses to chemotherapy. More specifically, the authors used supervised support vector ML to determine the gene sets whose expression was related to the specific tumor cell line GI50 . The authors discovered that specific genes and functional pathways can be used to distinguish which tumor cell lines are sensitive to chemotherapy drugs and which tumor cell lines are resistant to chemotherapy drugs. They tested their algorithm on bladder, ovarian and colorectal cancer patient data from The Cancer Genome Atlas (TCGA) in order to determine the response of tumor cell line GI50 to three chemotherapy drugs (cisplatin, carboplatin and oxaliplatin). Through experimental results, the authors found that for cisplatin, their algorithm was 71.0\% accurate at predicting disease recurrence and 59\% accurate at predicting remission. In the case of carboplatin, their algorithm was 60.2\% accurate at predicting disease recurrence and 61\% accurate at predicting remission. Finally, for oxaliplatin, their algorithm was 54.5\% accurate at predicting disease recurrence and 72\% accurate at predicting remission. Furthermore, in patients who used cisplatin and had a specific genetic signature, the algorithm was able to predict 100\% of recurrence in non-smoking bladder cancer patients and 79\% recurrence in smokers. 
\subsubsection{Research Opportunities:}
\indent Many research opportunities still exist in applying ML for drug response prediction. One such opportunity is extending existing models to to predict the drug responses of cancer patients who are receiving emerging immuno- and other targeted gene therapies. This will validate the comprehensiveness and generality of the considered frameworks. Another potential research opportunity is to build more comprehensive models by using more drug features (such as concentration, SMILES strings, molecular graph convolution and atomic convolution). This again will help extract more information and potentially uncover more correlations and inter-dependencies that can make the models more robust and accurate. A third opportunity is to investigate other methods and techniques including semi-supervised learning methods to encode molecular features with external gene expression and other types of data. This particularly would be helpful given that access to labeled data is not always possible. Therefore, having semi-supervised based ML models can help healthcare professionals gain insight from labeled data and apply it to the unlabeled data that they have. Last but not least, researchers should also investigate ways to adapt existing models to other drugs, cancer types, and diseases. This is essential as it would provide one adaptive system that can help healthcare professionals from different specializations make use of the available data. 
\subsection{Diabetes Research}
\subsubsection{Challenge Description:}
\indent As mentioned before, many modern medical organizations use electronic health records (EHRs) to store the medical data of patients. The large amounts of data present in EHRs can be a valuable source for researching diabetes mellitus (DM). \citet{s45} discuss what DM is and why it is a medical concern. DM is a group of metabolic disorders that are mainly caused by abnormal insulin secretion and/or action. Abnormal insulin secretion can result in a patient’s body not producing enough insulin which causes the patient’s metabolism of carbohydrates, fat and proteins to be impaired, which in turn results in elevated blood glucose levels (hyperglycaemia).\\
\indent There are two major clinical types of DM, type 1 diabetes (T1D) and type 2 diabetes (T2D). T1D is linked to the auto-immunological destruction of the Langerhans islets; whereas T2D is linked to lifestyle, little physical activity, poor dietary habits and heredity. The main treatment for T1D is insulin administration which can applied to T2D patients. However, the main treatment for T2D is improved diet, weight loss, exercise and oral medication. DM affects more than 200 million people worldwide, with 10\% of those affected with T1D and 90\% affected with T2D. DM possess a health threat as chronic hyperglycaemia results in several complications, including diabetic nephropathy, retinopathy, neuropathy, diabetic coma and cardiovascular disease. 
\subsubsection{Previous Works:} 
\indent DM has a high mortality and morbidity rate, therefore, detecting and treating DM is of high interest to the medical community as well as those who may or already do suffer from DM \citep{s45}. In recent years, researchers have been able to apply ML algorithms to the data of patients with DM in order to improve the methods of detecting and treating DM. \\
\indent Hemoglobin is a substance in red blood cells that carries oxygen to tissues. However, it can also attach to sugar in the blood and form a substance called glycated hemoglobin (HbA1c) \citep{s48}. A patient’s HbA1c level can be checked in order to determine if they have T2D. Alternatively, a patient’s HbA1c level along with their fasting blood glucose level and oral glucose tolerance test results can be used to determine if they have T2D \citep{s46}. Currently, to diagnosis a patient with T2D their HbA1c value must be at or above 6.5\% . However, studies have shown that the cut-off value of 6.5\% leads to inconsistencies in the diagnosis of T2D. Hence, using HbA1c with a 6.5\% cut-off value as a single marker for T2D may lead to undiagnosed cases of diabetes.\\ 
\indent \citet{s46} applied ML algorithms to the data of 840 patients from the Diabetes Health screening (DiabHealth) in order to identify an optimal cut-off value for HbA1c and to identify whether additional biomarkers could be used along with HbA1c to increase the diagnosis of T2D. Then the authors used T2D as the class feature and generated a conventional DT using an information gain (IG) measure. Using this algorithm, the authors found that if an oxidative stress marker (8-OhdG) was included in the model along with HbA1c that the accuracy of detecting T2D at the 6.5\% HbA1c level increased from 78.71\% to 86.64\%. The authors also found that if interleukin-6 (IL-6) was included in the model along with HbA1c that the accuracy of detecting T2D increased from 78.71\% to 85.63\%. However, in this model, the optimal HbA1c range was between 5.73 and 6.22\% .\\ 
\indent \citet{s47} used ML to improve the treatment methods of T1D. Diabetics with T1D need to use the medication insulin in order to maintain normal blood sugar levels. Diabetics must self-administer multiple daily injections of insulin, both before meals and basally, in order to mimic the natural insulin secretion of the pancreas. Before diabetics administer these injections, they must prick their fingertip to draw blood that is placed in an electronic glucose meter that determines the amount of glucose in the patient’s blood \citep{s49,s50}. However, in recent years, an alternative form of therapy has become available. This alternative form of therapy is insulin pump therapy. In this therapy, injections are provided by continuous subcutaneous insulin infusion. The application of insulin by a machine allows diabetics to avoid multiple uncomfortable finger pricks and injections. Insulin that is taken at meal times is referred to as bolus insulin. Typically, bolus insulin doses are calculated by estimating carbohydrate intake and dividing this number by a fixed carbohydrate to insulin ratio, then adding a correction dose derived from the individual’s insulin sensitivity factor \citep{s47}. Although several algorithms have been developed to calculate bolus insulin dose \citep{s51,s52,s53,s54,s55,s56}, these algorithms have only been incorporated in commercially available insulin pumps and in some glucose meters \citep{s57}. However, these algorithms have not been adopted widely commercially due to economic risk, security issues and inertia to change, and the lack of ease of use \citep{s58}. Based on these challenges, \citet{s47} set out to create a more user-friendly bolus insulin calculating system. The authors used a decision support algorithm that incorporated Run-To-Run (R2R) control and case-based reasoning (CBR). They tested their algorithm via in-silico scenarios by using a simulator that emulated intra-subject insulin sensitivity variations and uncertainty in the capillarity measurements and carbohydrate intake. Via these simulations, the authors found that the CBR(R2R) algorithm significantly reduced the mean blood glucose level and completely eliminated hypoglycemia. When the authors compared the CBR(R2R) algorithm to a standalone (R2R only) version of the algorithm, they found that the CBR(R2R) algorithm performed better. The goal of the algorithm was to reduce blood glucose levels. Therefore, the CBR(R2R) algorithm performed better than the standalone R2R algorithm in both populations.\\
\begin{figure}[!htbp]
	\centering
	\includegraphics[scale=0.3]{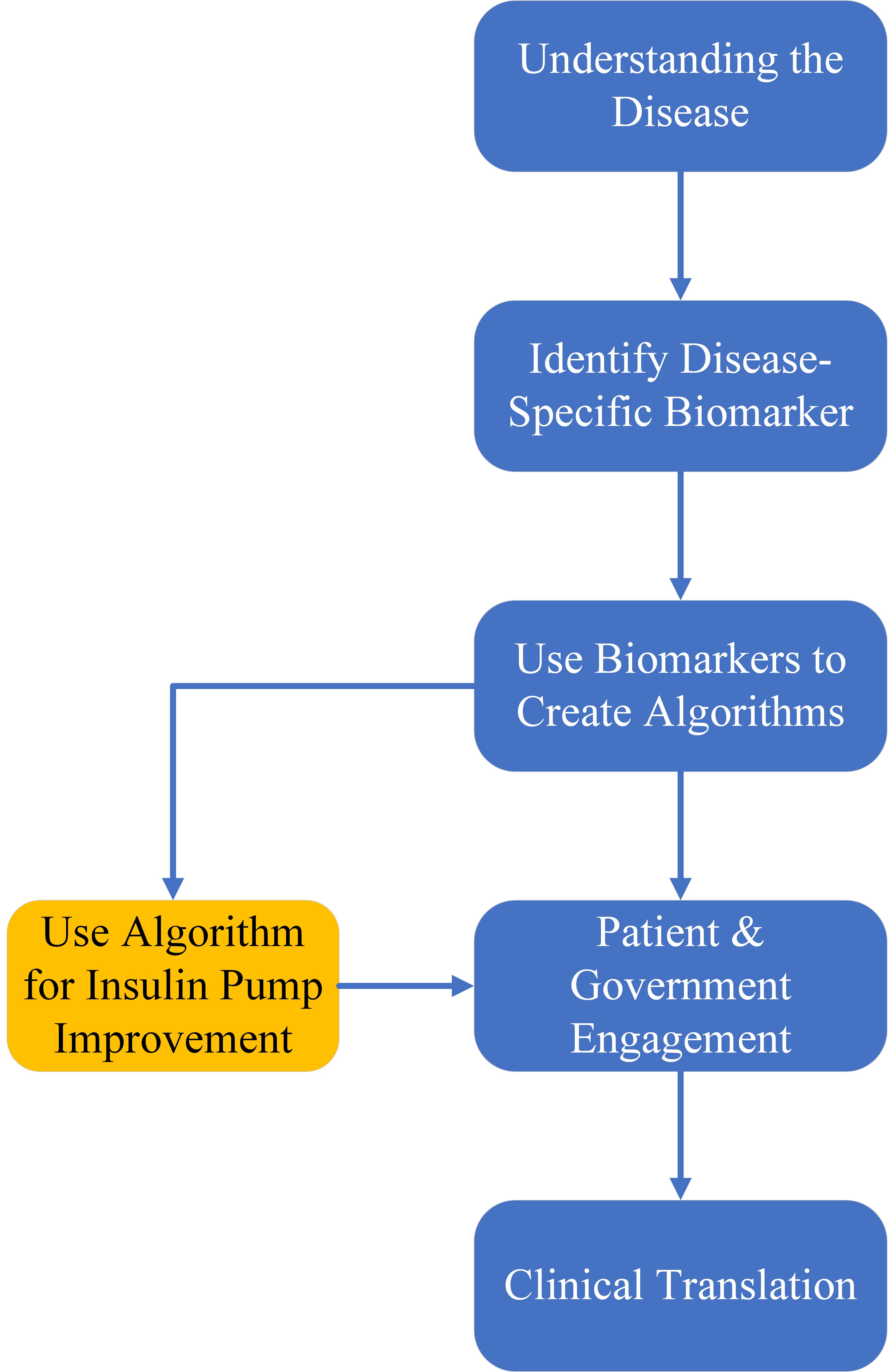}
	\caption{Potential Deployment of ML in Diabetes Research}
	\label{ML_in_healthcare}
\end{figure}
\subsubsection{Research Opportunities:}
\indent Despite the fact that ML has been used in diabetes research, more opportunities still exist. One suggestion is testing existing algorithms in the real-world via clinical trials. This is particularly important given that simulation environment tend to over-estimate the benefits of an intervention and may not always provide an accurate representation of the behavior of the body. Another opportunity worth exploring is investigating the performance of different ML classification models. This can help validate whether existing models have any bias. Therefore, it is important to compare the performance of different models to have a more accurate and sensitive model for insulin calculation. Figure \ref{ML_in_healthcare} provides a visualization of how these topics fit into a precision medicine framework.
\subsection{Retinopathy Detection Through Image Classification}
\subsubsection{Challenge Description:}
\indent Another healthcare-related area in which ML is playing a major role is for the detection of retinopathy. Retinopathy refers to the damage of the retina, the light-sensing inner part of the human eye \citep{retino_definition}. This can be due to multiple causes and diseases which can lead to partial or complete vision loss. There are different types of retinopathy including: retinopathy of prematurity, diabetic retinopathy, hypertensive retinopathy, and central serous retinopathy. Detecting the damage at an early stage is crucial to facilitate the treatment and slow down the loss vision process. To that end, multiple previous works proposed the use of ML algorithms and paradigms to accurately detect retinopathy. 
\subsubsection{Previous Works:} 
\indent \citet{retino1} proposed the use of ensemble ML models to perform early detection of diabetic retinopathy. More specifically, the authors proposed ensembles based on DT, adaBoost, Naive Bayes, K-NN, RF, and SVM to detect this disease using features extracted from retinal images such as diameter of optic disk, lesion specific, and image level features. Their experiments showed that the proposed models achieved a detection accuracy of up to 94\%.\\ 
\indent Similarly, \citet{retino2} also proposed the use of ensemble models to detect ABCA4-Related Retinopathy. The authors used high-dimensional microstructural eye image's dataset and extracted multiple features. The authors then developed different ensemble learning models based on K-NN, RF, SVM, and eXtreme Gradient Boosting (XGBoost). Their experimental results showed that the proposed model achieved detection accuracies ranging between 86\%-93\%.\\
\indent \citet{retino3} also proposed the use of ensemble-based ML models to detect diabetic retinopathy. In their work, the authors considered multiple classifiers including RF, DT, Adaboost, K-NN, and Logistic Regression (LR). These methods were applied to a diabetic retinopathy dataset that was normalized using the min-max method. Results showed that the proposed ensemble model outperformed the base models and achieved a detection accuracy above 80\%.\\
\indent On the other hand, \citet{retino4} proposed the combination of principal component analysis (PCA) and DNN for the early detection of diabetic retinopathy. The authors used a diabetes retinopathy dataset available at the UCI machine learning repository and normalized it using the Z-score technique. After normalization, PCA was applied to extract the most significant features. The reduced dataset was then given as an input to a DNN model for classification. The experimental results showed that the proposed DNN model achieved training accuracies between 72\%-82\% and testing accuracies between 68\%-79\%.\\
\indent Similarly, \citet{skb_h1} used DNN for the automated identification and grading system of diabetic retinopathy. The proposed system uses transfer learning and ensemble learning to detect the presence and severity of DR from fundus images. The authors' experimental results showed that their developed model has a high identification sensitivity of 97.5\% and a specificity of
97.7\%. On the other hand, the grading model achieved a sensitivity of 98.1\% and a specificity of 98.9\%.
\subsubsection{Research Opportunities:}
\indent Despite the promising results shown by ML models for retinopathy detection, further opportunities still exist. One such opportunity is developing optimized ML models. This is because many of the previous works in the literature only consider the default version of the classifiers. However, it is important to explore the impact of different optimization methods on the overall performance of the classifiers. Another opportunity is to consider different deep learning models such as convolutional neural networks (CNN) and recurrent neural networks (RNN). More specifically, CNNs and RNNs can be combined to extend the capabilities of traditional CNN models from the binary to multi-label image classification as illustrated in \citep{cnn_rnn_image,cnn_rnn_image2}. Such architectures have the potential to further improve the performance of deep learning models for early detection of retinopathy.
\subsection{Cancer Detection Through Image Classification}
\subsubsection{Challenge Description:}
\indent Another prominent area in which ML is used within the healthcare field is the detection of different types of cancer including breast cancer, prostate cancer, and lung cancer. This is often done by applying ML methods to images of the different organs or tissue suspected to have cancer \citep{cancer_ml}. Due to their success in image classification problems in general, ML models have been proposed in multiple research works from the literature to detect cancer based on tissue images.
\subsubsection{Previous Works:} 
\indent \citet{cancer1} investigated the performance of six different ML algorithms to detect breast cancer. More specifically, the author compared between linear regression, multi-layer perceptron (MLP), K-NN, softmax regression, and two variants of SVM algorithm. To evaluate the performance of these algorithms, the Wisconsin diagnostic breast cancer dataset was used which is composed of features extracted from digitized images of tests on breast mass. Experimental results showed that the detection accuracy ranged between 93\%-99\% with the MLP algorithm being the most accurate.\\
\indent Similarly, \citet{cancer2} also proposed the use of ML algorithms for breast cancer detection. However, the authors in this case developed a deep learning model, namely CNN, that was applied to digitized film mammograms from the Digital Database for Screening Mammography. Experimental results showed that the developed CNN model achieved accuracies between 63\%-99\% with the performance dependent on whether the CNN was pre-trained or not. Moreover, the developed model achieved an area under the curve (AUC) value reaching 0.88, illustrating its effectiveness and robustness in detecting breast cancer.\\
\indent On the other hand, \citet{cancer3} proposed the use of ML models for prostate cancer. To that end, the authors explored different ML algorithms such as Bayesian approach, SVM with multiple kernels, and DT. The authors also investigated different feature extraction strategies based on texture, morphological, scale invariant feature transform (SIFT), and elliptic Fourier descriptors (EFDs) features. Experimental results showed that the SVM classifier with RBF kernel achieved the highest accuracy ranging between 98\%-99\%.\\
\indent \citet{cancer4} proposed the use of ML algorithms to detect lung cancer based on computed tomography (CT) scan images. To that end, the authors proposed the use of a neural network-based model, namely the entropy degradation method (EDM), to detect cancerous images. The performance of the proposed model was explored using a high-resolution CT scan images provided by the National Cancer institute. Experimental results showed that the proposed model achieved a detection accuracy of 77.8\%, illustrating its effectiveness to detect small-cell lung cancer at an early stage.
\begin{table}[!h]
	\centering
	\caption{\label{ML_health_table} Challenges, Previous Works, and Research Opportunities within Healthcare Sector}
	\scalebox{0.9}{
	\begin{tabular}{|p{1.5cm}|p{4.5cm}|p{4.5cm}|}
		\hline
		Challenge & Previous Work & Research Opportunity  \\ \hline
		\multirow{4}{1.5cm}{Drug Response Prediction}&Applied (SVM)-based algorithm to predict the responses of individuals to a variety of standard-of-care chemotherapeutic drugs \citep{s40}& - Extend existing models to to predict the drug responses of cancer patients who are receiving emerging immuno- and other targeted gene therapies.\\ \cline{2-2} 
		&Developed a deep learning model to predict cell line response to a subset of drug pairs in the National Cancer Institute-ALMANAC database \citep{s42}&- Build more comprehensive models by using more drug features.\\ \cline{2-2} 
		&Used DNN and the genomic profiles of cancer tumors to predict the tumors’ responses to therapeutic drugs \citep{s43}& - Investigate other methods and techniques including semi-supervised learning methods on other types of data.\\ \cline{2-2} 
		&Used SVM to determine the gene sets whose expression was related to the specific tumor cell line GI50 \citep{s44}&- Investigate ways to adapt existing models to other drugs, cancer types, and diseases. \\ \hline
		\multirow{3}{1.5cm}{Diabetes Research}& Used conventional DT to identify an optimal cut-off value for HbA1c \citep{s46}& - Investigate the performance of different ML classification models to validate whether existing models have any bias.\\ \cline{2-2} 
		& Used a decision support algorithm to calculate the bolus insulin levels \citep{s47}&- Test existing algorithms in the real-world via clinical trials. \\ \hline
		\multirow{5}{1.5cm}{Retinopathy Detection Through Image Classification}& Used of ensemble ML models based on multiple algorithms to perform early detection of diabetic retinopathy\citep{retino1}& - Develop optimized ML models to further improve detection accuracy. \\ \cline{2-2} 
		& Used ensemble models to detect ABCA4-Related Retinopathy \citep{retino2}& - Consider different deep learning models and architectures such as CNN and RNN.\\ \cline{2-2}
		& Proposed ensemble-based ML models to detect diabetic retinopathy\citep{retino3}&- Use DNN techniques to build models that predict if a patient is diabetic or not. \\ \cline{2-2} 
		& Proposed the combination of PCA and DNN for the early detection of diabetic retinopathy \citep{retino4}&\\ \cline{2-2}
		& Used DNN for the automated identification and grading system of diabetic retinopathy \citep{skb_h1}& \\ \hline
		\multirow{4}{1.5cm}{Cancer Detection Through Image Classification}& Investigated the performance of six different ML algorithms to detect breast cancer \citep{cancer1} & - Develop optimized ML models to further improve detection accuracy. \\ \cline{2-2} 
		&Proposed a deep learning CNN model for breast cancer detection \citep{cancer2}& - Apply different types of ensemble models that can combine multiple ML classifiers to improve their effectiveness and robustness. \\ \cline{2-2}
		& Explored different ML classifiers to detect prostate cancer \citep{cancer3} &- Consider different deep learning models and architectures such as RNN. \\ \cline{2-2} 
		&Proposed the EDM (a neural network-based model) model to detect small-cell lung cancer \citep{cancer4} &- Explore different algorithms to improve the feature extraction and selection process. \\ \hline
	\end{tabular}}
\end{table}
\subsubsection{Research Opportunities:}
\indent As shown above, ML algorithms have been successfully applied to detect different types of cancer. However, there still exists further opportunities to improve the detection performance. One such opportunity is to consider different hyper-parameter optimization methods to improve the performance of the ML models. Another potential opportunity is applying different types of ensemble models that can combine multiple ML classifiers to improve their effectiveness and robustness. A third opportunity is studying other deep learning techniques and architectures such as the combined CNN-RNN models to investigate their effectiveness in detecting cancer. An additional opportunity is to improve the feature extraction and selection algorithms from digital images. This is crucial given the fact that this process acts as the input to the ML model development stage. Therefore, it is important to extract and select relevant and high-quality features to be fed to the ML models under consideration.\\
\indent Similar to the previous section on education, Table \ref{ML_health_table} summarizes some of the challenges facing the healthcare sector, lists some of the previous works, and presents the different research opportunities. 
\section{Network Security}\label{network_sec}
\indent Turning to a different sector, ML can also be beneficial in network security. Cisco Systems, Inc., an American multinational technology conglomerate who specializes in information technology, networking, and cybersecurity solutions, defines network security as any activity designed to protect the usability and integrity of a network and data \citep{s59}. According to Cisco Systems, Inc., network security allows authorized users to assess a network while preventing outside threats from entering or spreading on a network \citep{se,se2}. Cisco Systems, Inc. lists fourteen types of network security. However, this section will focus on Intrusion Detection Systems (IDS). IDSs analyze and monitor network traffic in order to determine if the network traffic patterns show normal activity or if there are signs of malicious activity \citep{s60,s61}. More specifically, this section will discuss how ML can be used to improve network intrusion detection systems (NIDS) in general, how to better detect Botnets, and how to improve NIDS in vehicles.
\subsection{Network Intrusion Detection Systems}
\subsubsection{Challenge Description:}
\indent A Network Intrusion Detection System (NIDS) helps system administrators to detect network security breaches in their organizations \citep{s60,sf}. NIDSs are classified based on the style of detection that they use. Misuse-detection NIDSs use precise descriptions of known malicious behavior. Anomaly-detection NIDSs flag deviations from normal activity. Specification-based NIDSs define allowed types of activity and flag any other activity as forbidden. Behavioral detection NIDSs analyze patterns of activity and surrounding context to find secondary evidence of attacks. Although there are many types of NIDS, misuse-detection and anomaly-detection NIDSs are the most common \citep{s61}.\\
\indent Misuse-detection NIDSs can also be referred to as signature (misuse) based NIDS (SNIDS). In SNIDS, attack signatures are pre-installed in the NIDS and pattern matching is then performed between network traffic and the installed signatures. When a mismatch is found, it is considered an intrusion. There are advantages and disadvantages to both SNIDS and anomaly-detection NIDS (ADNIDS). SNIDSs are effective in the detection of known attacks and show high detection accuracy with less false-alarm rates, but is ineffective at detecting unknown or new attacks whose signatures have not been installed on the IDS. On the other hand, ADNIDSs are the better option for the detection of unknown and new attacks, but produce high false-positive rates. The current deployment framework and usage patters of NIDSs makes it hard for these systems to be efficient and flexible when considering unknown future attacks \citep{s60}. 
\subsubsection{Previous Works:} 
\indent \citet{s60} propose a solution to the challenges of using NIDS to detect known and unknown future attacks. The authors’ solution involved using a deep learning approach known as Self-Taught Learning (STL). The authors verified their method on the benchmark intrusion dataset NSL-KDD. This dataset is an improved version of the former benchmark intrusion dataset KDD Cup 99. The authors present various metrics related to their algorithm, including accuracy, precision, recall, and f-measure values. Experimental results showed that the authors’ algorithm achieved a classification accuracy rate above 98\%. \\
\indent \citet{sc} proposed using Bayesian optimization to hyper-tune the parameters of different supervised ML algorithms for anomaly-based IDSs. More specifically, they tune the parameters of SVM, Random Forest (RF), and k-NN algorithms. Then, the authors evaluated the performance of the regular and optimized version of these classifiers in terms of accuracy, precision, and false alarm rate. Their experimental results showed that the proposed framework achieved a high accuracy rate and precision, and a low-false alarm rate and recall.\\
\indent \citet{injadat_ch7} extended the work by proposing a novel multi-stage optimized ML-based NIDS framework. The goal of this framework is to reduce the computational complexity and maintain the detection performance. The performance of the proposed framework was measured using two state-of-the-art intrusion detection datasets, the CICIDS 2017 and the UNSW-NB 2015 datasets. Experimental results showed that the proposed model significantly reduced the required training sample size and feature set size. More specifically, the model reduced the training sample size by 74\% and the feature size by up to 50\%. Moreover, hyper-parameter optimization helped improve the model performance with the detection accuracies being over 99\% for both datasets. This represents an improvement of 1-2\% in terms of accuracy and 1-2\% in false alarm rate when compared to other works from the literature.\\
\indent \citet{sg} proposed an ensemble feature selection and an anomaly detection method for network intrusion detection. The proposed framework combined unsupervised and supervised ML techniques to classify network traffic and identify previously unseen attack patterns. To that end, the authors used three different feature selection techniques that identified 8 common and representative features. Moreover, the authors adopted k-Means clustering to segregate the training instances and developed the classification model accordingly. Their experimental results showed that the proposed framework was effective in detecting previously unseen attack patterns in comparison to the traditional classification approaches.\\
\indent \citet{skb_ids1} proposed the use of an SVM with augmented features for their intrusion detection framework. More specifically, the author used the logarithm marginal density ratios transformation to get better-quality features. Using the NSL-KDD dataset, their experiments showed that the proposed framework achieved better performance in terms of accuracy, detection rate, false alarm rate and efficiency.
\subsubsection{Research Opportunities:}
\indent Although the use of ML for network intrusion detection is popular, it still requires further research. One potential opportunity is to study the performance of more complex models such as bagging ensemble models or deep learning models. This is particularly important for real-time or near real-time network intrusion detection. Another opportunity is to study the impact of different optimization models and techniques in enhancing the current intrusion detection frameworks and models. A third research opportunity is to consider time-series analysis techniques to identify and detect temporal-based anomalies and intrusion attempts. This is crucial given that many attacks such as denial-of-service (DoS) attacks span a period of time rather than being instantaneous. Another opportunity is to investigate the performance of reinforcement learning and transfer learning techniques in IDSs. This is based on the fact that such techniques have the potential to make the IDSs more flexible and effective. Although there have been some works that have considered the use deep learning models and time series analysis such as the work by \citet{DL_IDS}, this should be extended to other network security problems and not just for IDSs.
\subsection{Botnets Detection}
\subsubsection{Challenge Description:}
\indent The term botnet refers to a network of computers (bots) which have been compromised by an attacker (aka botmaster) who has installed malicious software on the network via an attacking technique such as trojan horses, worms and viruses. Botmasters often choose to attack computer networks that contain many computers due to the large amounts of bandwidth and powerful computing capabilities available for such networks. Once the botmaster has control of a network, they use the network to initiate various malicious activities such as email spam, distributed denial-of-service (DDOS) attacks, password cracking, and key logging \citep{s62,icm_dns,icm_iot}. \\
\indent \citet{s63} divided the botnet life-cycle into three phases: 1. formation, 2. Command and control (C\&C), 3. botnet application phase. In the formation phase, the botmaster infects other machines on the Internet, turning them into bots on the botnet. In the C\&C phase, bots receive instructions from the bot master. During the botnet application phase, the bots carry out malicious activities based on the instructions of the botmaster. Although some bots might be detected and removed from the botnet, the botmaster will continue to probe the botnet in a stealthy manner for information about active bots and will plan to form a new botnet \citep{s64}.\\
\indent One common type of botnet is the Internet relay chat (IRC) botnet. This botnet uses IRC to facilitate command and control (C\&C) communication between bots and botmasters. IRC botnets can connect to one or more servers, making it easy for the botmaster to execute commands. However, IRC botnets can be stopped by shutting-down the IRC botnet’s C\&C server. Once attackers realized this central flaw of IRC botnets, they began to utilize peer-to-peer (P2P) botnets. In a P2P botnet, there is no centralized server and bots are connected to each other topologically and act as both C\&C server and client. Therefore, even if a P2P botnet loses some of its bots, its communication will not be disrupted. According to \citet{s62}, botnets have become one of the most significant threats to the Internet.
\subsubsection{Previous Works:} 
\indent \citet{s64a} proposed a deep learning-based model to detect botnet activity within IoT devices and networks. More specifically, the authors developed a Bidirectional Long Short Term Memory based Recurrent Neural Network (BLSTM-RNN). The detection model was compared to a default LSTM-RNN. The performance was evaluated using the accuracy and loss metrics. Experimental results demonstrated that the proposed model achieved a detection accuracy ranging between 92\%-99\%, highlighting the effectiveness of RNN-based models for botnet detection.\\
\indent \citet{s66} investigated the ability of three ML algorithms (RF, LR, and SVM) to effectively select features to use in botnet detection during network flow analysis. More specifically, the authors investigated three different feature selection methods. This included Lasso linear regression models, Recursive Feature Elimination (RFE), and tree-based feature selection, along with three different classifiers (LR, NB, and RF). The authors found that when the meta-classifier RF was applied on the features selected by RF that the model was nearly 99.9\% accurate, making it the most accurate model that was tested. This model almost achieved perfect classification accuracy for identifying botnet and normal traffic.\\
\indent \citet{s67} also discussed using ML to detect botnets. According to the authors, in the past, signature-based and anomaly-based intrusion detection systems (IDS) were used to detect botnets. However, as the speed of the Internet has increased, these methods are no longer as effective. The authors proposed a method that uses conversation-based network traffic analysis and supervised ML to identify malicious botnet traffic . The authors showed that their approach outperformed other approaches which are based on network flow analysis. More specifically, the authors’ model resulted in a 13.2\% decrease in the false positive rate of botnet traffic detection. Furthermore, it was shown that the RF algorithm had a high detection accuracy (93.6\%) and a low false positive rate (0.3\%).
\subsubsection{Research Opportunities:}
\indent As mentioned earlier, there are still many research opportunities in the usage of ML for botnet detection that are worth exploring. One such opportunity is investigating the use of hybrid ML models to see if they can satisfy all the requirements of their proposed online botnet detection framework. Another potential opportunity is to consider non-numerical features as part of any botnet detection models since such features may contain valuable information. A third opportunity again is studying the impact of different optimization models and techniques on the performance of current botnet detection models.
\subsection{Intrusion Detection in Vehicles}
\subsubsection{Challenge Description:} 
\indent In recent years, the conventional mechanical controlling parts in cars have largely been replaced by Electronics Control Units (ECUs) \citep{s69}. ECUs are computing devices that are used for controlling and monitoring the subsystems of a vehicle for energy efficiency enhancement, and noise and vibration reduction \citep{s68,AM_V2X1}. The use of computing devices in vehicles has led to the use of automotive networking services such as Vehicle-to-Vehicle (V2V) and Vehicle-to-Infrastructure (V2I) services. V2V automotive networking services require computing devices to perform intra-vehicular communication, while V2I automotive networking services require computing devices to perform inter-vehicular communication \citep{AM_V2X2}.\\
\indent One standard communication protocol for in-vehicle network communication is Controller Area Network (CAN). CAN connects sensors and actuators with ECUs \citep{sd}. Important information such as diagnostic, informative, and controlling data is delivered through a CAN bus and it is important that this information is secured in order to keep the driver safe. However, whenever networks are used, there is a potential for significant security concerns. For in-vehicular networks there are several security flaws. For example, ECUs can obtain any ECU-to-ECU broadcasting messages in the same bus, but they are unable to identify a sender \citep{s68}.
\subsubsection{Previous Works:} 
\indent Based on their concerns about the security issues of in-vehicular networks, especially the CAN bus component, \citet{s68} created an intrusion detection system that uses a deep neural network (DNN). The authors’ DNN was able to more accurately detect intrusions than a traditional ANN. According to the authors, this increased accuracy is due to the deep learning framework, which allows for the initialization of parameters through the unsupervised pre-training of deep belief networks (DBN). Finally, using experimental results, the authors showed that their algorithm can provide a real-time response to an attack with a detection ratio average of 98\%.\\
\indent In a similar manner, \citet{sd} proposed an IDS for autonomous and connected vehicles using DT structures. The goal of the IDS is to detect network attacks within the vehicle and external to it. Experimental results showed that the authors' proposed framework improved the detection accuracy, detection rate, and F1 score by close to 2-3\% and achieved lower false alarm rate than other traditional methods proposed in the literature. Moreover, the developed IDS detected various attacks. The proposed model achieved an accuracy of 100\% and 99.86\% on the CAN intrusion and CICIDS2017 data sets. Additionally, it reduced the computational time by 73.7\% to 325.6s and by 38.6\% to 2774.8s, respectively.\\
\indent \citet{s75} proposed a deep Learning-based intrusion detection model composed of a combination of CNN and LSTM to detect malware traffic for on board units. The proposed model is fed the raw traffic instead of the human-extracted private information features. The performance of the proposed model was compared with previous methods on a public dataset and a simulated real-life VANET dataset. Experimental results showed that the proposed model outperformed the other methods by achieving a precision value between 95\%-99\% and an F1-score between 0.92-0.99.
\begin{figure}[!bt]
	\centering
	\includegraphics[scale=0.3]{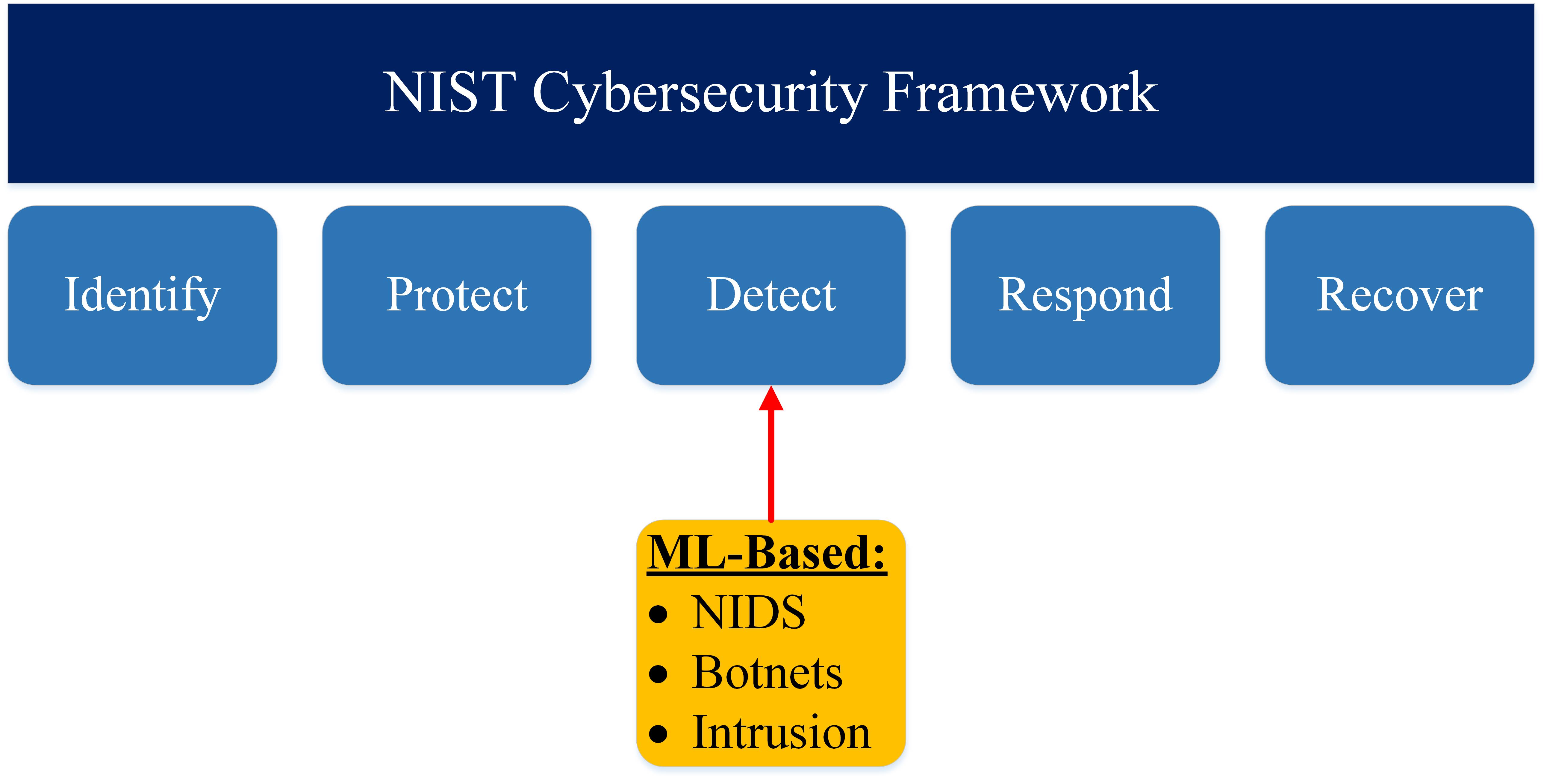}
	\caption{Potential Deployment of ML in Network Security}
	\label{ML_in_security}
\end{figure}
\begin{table}[!tbp]
	\centering
	\caption{\label{ML_security_table1} Challenges, Previous Works, and Research Opportunities within Network Security Field}
	\scalebox{0.88}{
		\begin{tabular}{|p{1.5cm}|p{4.5cm}|p{4.5cm}|}
			\hline
			Challenge & Previous Work & Research Opportunity  \\ \hline
			\multirow{5}{1.5cm}{Network Intrusion Detection Systems}&Used a deep learning approach to detect network intrusions \citep{s60}& - Study the performance of more complex models such as bagging ensemble models, deep learning models, reinforcement learning, and transfer learning. \\ \cline{2-2}
			&Used Bayesian optimization to hyper-tune the parameters of three classification algorithms for anomaly-based IDSs \citep{sc}& - Study the impact of different optimization models and techniques in enhancing the current intrusion detection frameworks and models.\\ \cline{2-2}
			&Proposed a novel multi-stage optimized ML-based NIDS framework that reduced the computational complexity while maintaining its detection performance \citep{injadat_ch7}& - Consider time-series analysis techniques to identify and detect temporal-based anomalies and intrusion attempts.\\ \cline{2-2}
			&Proposed an ensemble feature selection and an anomaly detection method for network intrusion detection \citep{sg}& - Explore the performance of NIDS using recent datasets such as CICIDS2017, CSE-CIC-IDS2018 and Kyoto 2006+. \\ \cline{2-2}
			&Used SVM with augmented features for their intrusion detection framework \citep{skb_ids1}&\\ \hline
			\multirow{3}{1.5cm}{Botnets Detection}& Proposed a BLSTM-RNN model to detect botnets \citep{s64a}& - Investigate the use of hybrid ML models to see if they can satisfy all the requirements of their proposed online botnet detection framework.\\ \cline{2-2}
			&Used three ML algorithms to perform botnet detection during network flow analysis \citep{s66}.& - Study the impact of different optimization models and techniques on current botnet detection frameworks and models.\\ \cline{2-2}
			&Used conversation-based network traffic analysis and supervised ML to identify malicious botnet traffic \citep{s67}& - Consider non-numerical features as part of any botnet detection models since such features may contain valuable information. \\ \hline
			\multirow{3}{1.5cm}{Intrusion Detection in Vehicles}&Used a deep neural network (DNN) for intrusion detection in vehicles \citep{s68}& -  Explore the impact of different optimization techniques and meta-heuristics to tune the hyper-parameters of existing IDS models. \\ \cline{2-2}
			&Proposed a DT-based IDS for autonomous/connected vehicles \citep{sd}&- Develop more complex and hybrid ML systems that can detect both the known and unknown attacks in vehicular networks. \\ \cline{2-2}
			&Proposed a deep Learning-based intrusion detection model composed of a combination of CNN and LSTM to detect malware traffic for on board units \citep{s75}& \\ \hline
	\end{tabular}}
\end{table}
\subsubsection{Research Opportunities:}
\indent There is still ample research opportunities to integrate ML as part of IDS systems for vehicular networks. For example, it is worth exploring the impact of different optimization techniques and meta-heuristics such as particle swarm optimization and Baysian optimization to tune the hyper-parameters of existing IDS models \citep{hyper1}. This should be done in order to improve the overall performance of such models. Another potential research opportunity is developing more complex and hybrid ML systems that can detect both the known and unknown attacks in vehicular networks. This is particularly important since more novel attacks are being introduced that are targeting autonomous and connected vehicles. \\
\indent Table \ref{ML_security_table1} summarizes the previously discussed challenges and present some of the literature work that has been conducted within this field. Moreover, they also list some of the potential research opportunities in which ML can play a role. Also, Figure \ref{ML_in_security} provides a visualization of how Network Intrusion Detection System (NIDS), Detecting Botnets, and Intrusion Detection in Vehicles fit into the Detect level of the NIST’s Cybersecurity Framework.  
\section{Banking \& Finance}\label{buss_finance}
\indent Moving on from network security, ML also has application in the sectors of banking and finance. After the financial crises of the 1980’s and 90’s, risk assessment of financial intermediaries became a hot topic. ``A financial intermediary is an entity that acts as the middleman between two parties in a financial transaction, such as a commercial bank, investment banks, mutual funds and pension funds'' \citep{s77}. Researchers such as \citet{s76} believe that ML algorithms can be used to predict individual risk in the credit portfolios of institutions. In turn, this will help in determining who will and will not repay various forms of credit (e.g., loans, mortgages, and credit cards). Khandani \textit{et al.} echo this sentiment as they discuss the importance of using ``hard'' information (e.g., characteristics contained in consumer credit files collected by credit bureau agencies) to determine the creditworthiness of consumers \citep{s78}. In the past, human discretion has been used to determine the creditworthiness of consumers. However, ML offers a way to determine the creditworthiness of consumers based on vast amounts of hard information. This section will discuss how ML can be used to assess the credit risk of potential borrowers, predict if borrowers will go bankrupt, and predict currency crises. 
\subsection{Credit Risk Assessment}
\subsubsection{Challenge Description:}
\indent With the increased dependency on mortgages and banks for financial support, credit risk assessment has garnered significant interest from both practitioners and researchers. This is especially crucial for financial institutions to be able to differentiate between ``good'' and ``bad'' applicants to minimize their risk \citep{s78_a}. This applies to both individual applicants as well as small-medium enterprise (SMEs) applicants \citep{s78_b}. Multiple factors are typically considered when using traditional assessment systems \citep{skb_f2}. However, an applicant's dynamic transaction history, an important indicator of the applicant's trustworthiness and creditworthiness, is often not considered. Additionally, in the case of SMEs, the enterprise's ``self-oriented'' factor and ``supply chain finance-oriented'' factor are often neglected when assessing credit risk. Therefore, it is important for any credit assessment system to consider multiple factors to better utilize available resources. \\
\subsubsection{Previous Works:}
\indent \citet{s78_a} proposed the combined use of unsupervised and supervised ML models to assess the credit risk of individuals. More specifically, the authors explored two different clustering models, namely k-means and the self-organizing map (SOM) in addition to seven potential supervised classification models including LR, DT, gradient boosting decision tree (GBDT), RF, SVM, K-NN, and artificial neural networks (ANN). The authors studied the performance of their proposed models using three datasets from China, Germany, and Australia respectively. Experimental results showed that the detection accuracy of the potential models ranged between 81\%-91\% for the Chinese dataset, between 64\%-79\% for the German dataset, and 64\%-86\% for the Australian dataset.\\
\indent \citet{s78_b} proposed a hybrid ensemble ML model to assess the credit risk of SMEs in supply chain finance. The authors integrated two ensemble learning models, namely the random subspace (RS) model and the multi-boosting model based on DT algorithm to improve the performance of the credit risk assessment process. The performance of the proposed model was evaluated on a Chinese dataset collected between 31 March 2014 and 31 December 2015. Experimental results showed that the assessment accuracy ranged between 67\%-84\% with the hybrid RS-multi-boosting model achieving the highest accuracy.\\
\indent On the other hand, \citet{s78_c} proposed a deep learning model for SME credit risk assessment. More specifically, the authors proposed a DBN composed of the Restricted Botlzman Machine (RBM) and Softmax classifier to predict the credit risk of SMEs working in the online supply chain space. The authors evaluated the performance of their proposed model using three different datasets and was compared to the performance of SVM and LR models. Experimental results showed that the proposed DBN achieved the highest accuracy of 96\% compared to 82\% and 87\% for the LR and SVM models respectively.\\
\subsubsection{Research Opportunities:}
\indent Despite the literature showing that using ML has great potential for credit risk prediction, there are still research opportunities in this field. One potential opportunity to explore is optimizing the hyper-parameters of the ML models considered. As shown in the previous works, most of the proposed models only consider default parameters without any attempt to optimize them, which may result in reduced performance. Another potential opportunity is exploring the performance of different models that can make short, medium, and long term risk prediction rather than just on the short term. In a similar manner to the work in \citep{GAN_credit_assessment}, a third opportunity is to consider other deep learning models and architectures such as CNN, RNN, DBNs, and Generative Adversarial Networks (GANs) to investigate the improvement in the credit assessment accuracy. 
\subsection{Bankruptcy Prediction}
\subsubsection{Challenge Description:}
\indent When selecting potential clients one aspect of their amount of credit risk is the probability that they will go bankrupt. Quantitative risk management systems, which are based on ML models, can provide financial institutions with early warning signs of clients whose potential business may fail \citep{s79}. Such failure can result in bankruptcy and the client defaulting on their bank payments. In turn, this can have a devastating impact on the firm owner, society, and the country's overall economy \citep{s79_a}. This would force governments to increase their rescue plans in order to maintain the economic growth of the country which is a challenging task in itself. Prior work has used linear probability and multivariate conditional probability models, the recursive partitioning algorithm, artificial intelligence, multi-criteria decision making, and mathematical programming in order to predict a person’s amount of credit risk. However, the performance of the previously proposed models heavily depends on the features and data collected.
\subsubsection{Previous Works:} 
\indent \citet{s80} discuss how the financial sustainability of a company can maintain the soundness of the state and society. They further discuss how the sustainability of financial institutions is directly dependent on the financial sustainability of the bank’s borrowers. Hence, it is important for financial institutions to evaluate the sustainability of their borrowers, which is often done with the corporate financial distress prediction model. The authors propose a novel hybrid SVM model that uses globally optimized SVMs (GOSVM) and the genetic algorithm (GA) to predict potentially distressed burrowers. GOSVM optimizes feature selection, instance selection, and kernel parameters; while GA simultaneously optimizes multiple heterogeneous design factors of SVMs. The authors trained and tested their model on real-world data from H commercial bank in Korea. The authors randomly chose 1,548 heavy industry companies, 774 of which had filed for financial distress between 1999 and 2002, and 774 which were non-bankrupt in this same time period. Experimental results showed that the proposed GOSVM model outperformed both non-SVM based models and other SVM-based models at accurately predicting financial distress during the hold-out phase. Based on these results, the authors concluded that their model improves the prediction accuracy of conventional SVMs.\\
\indent Similarly, \citet{s80_a} proposed the use of different ML models to predict bankruptcy and default events of companies and institutions. More specifically, the authors studied four models, namely SVM, DT bagging, DT boosting, and RF models in comparison with other traditional models such as LR and ANN. Experimental results showed that the proposed models achieved higher prediction accuracy during both the training and testing stages. In particular, the bagging, boosting, and RF models all achieved a training accuracy above 96\% and a testing accuracy between 86\%-87\% as compared to the LR and ANN methods which achieved a training accuracy between 82\%-84\% and a testing accuracy between 72\%-76\%. \\
\indent \citet{s80_b} also proposed the use of ensemble learning models in combination with feature selection as part of their bankruptcy prediction models. To that end, the authors investigated two feature selection methods, namely information gain and genetic algorithm. The authors also explored six ML models including LR, NB, ANN, DT, SVM, and K-NN. Experimental results illustrated that the bagging ensemble models achieved better performance when compared to the single classifiers by having a lower false positive rate. Moreover, the results also showed that genetic algorithm outperformed the information gain algorithm for feature selection as it allowed the classifiers to achieve better performance.\\
\indent On the other hand, \citet{s80_c} proposed the use of deep learning models to predict bankruptcy based on textual data in conjunction with accounting-based ratio and market-based variables. In particular, the authors proposed a CNN-based model with word embedding as part of their bankruptcy prediction model. To evaluate the performance of their proposed models, the authors used a dataset consisting of 11,827 firms and 94,994 firm-years collected from Compustat North America, Center for Research in Security Prices (CRSP), and the Securities Exchange Commission (SEC). Experimental results showed that the proposed model outperformed other traditional models such as LR, RF, and SVM by achieving higher prediction accuracies.  
\subsubsection{Research Opportunities:}
\begin{figure}[!t]
	\centering
	\includegraphics[scale=0.35]{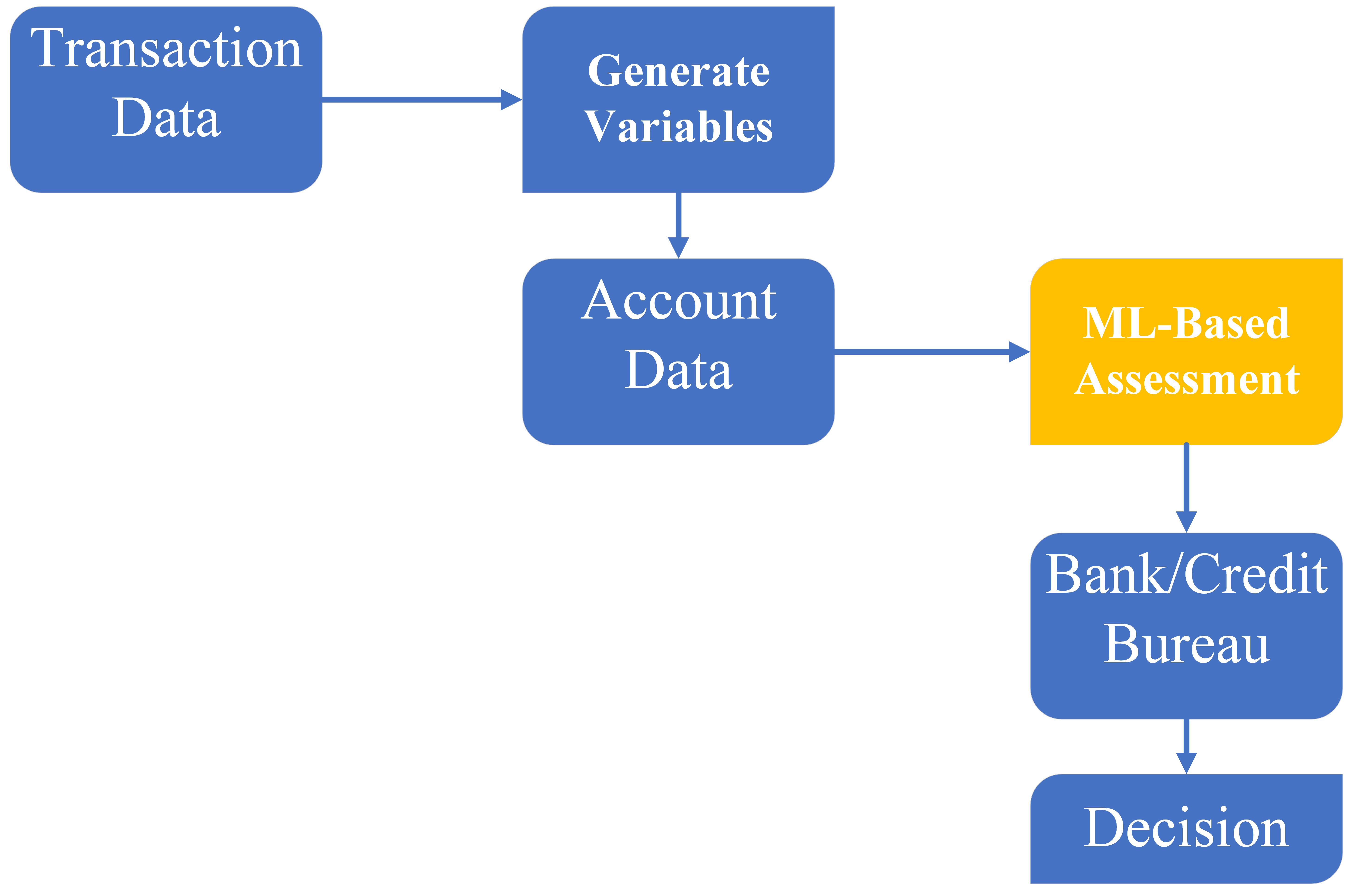}
	\caption{Potential ML-based Credit Risk and Bankruptcy Assessment Framework}
	\label{ML_in_business} 
\end{figure}

\indent Again, there are still many research opportunities that would benefit from ML to better predict bankruptcy. For example, one open area is studying the impact of other ML models and kernels in performing the prediction. This is based on the fact that most previous work only focused on a single kernel or a single ML model. Another research opportunity that should be considered is investigating the performance of different optimization models and meta-heuristics such as simulated annealing, tabu search, or particle swarm optimization to study the potential trade-off between performance improvement and computational complexity. A third research opportunity is to explore other deep learning models and architectures such as RNN, DBN, and GANs to investigate their performance in comparison with CNN models proposed in the literature. Figure \ref{ML_in_business} provides a potential ML-based credit risk and bankruptcy assessment framework that can be deployed by banks and financial institutions.
\subsection{Currency Crises Prediction}
\subsubsection{Challenge Description:}
\indent In the 90s, many countries suffered from a currency crisis wherein the value of their currency became unstable. Europe experienced a currency crisis in 1992, Mexico in 1994, Asia in 1997-98, and Russia in 1998 \citep{s81}. Therefore, the interest in developing such systems increased in the aftermath of the 2008-09 global financial crisis \citep{s81_a,s81_aa}. The interest stems from the fact that a currency crisis can damage the world economy. Hence, it would be beneficial to create an early warning system in order to prevent or at least to manage such events, particularly given the serious socio-economic impact that such events can have. To that end, ML can play a major role as part of such early systems given their ability to act as accurate prediction models and their promising performance in other business-focused applications.  
\subsubsection{Previous Works:} 
\indent \citet{s81_a} proposed the use of ML models to predict crises in different currencies. In particular, the author proposed the use of SVM model as it can overcome many of the limitations of traditional crises prediction approaches including data non-linearity and the variance-bias trade-off. To that end, the author applied the SVM model to data collected between 1996 and 2014. Experimental results showed that the proposed model accuracy predicted the majority of crises within that period from 17 emerging markets, thus illustrating its potential as a valuable tool for economists to use.\\
\indent On the other hand, \citet{s81_b} proposed the combination of RF and wavelet transformation to predict currency crises. More specifically, the authors proposed the use of discrete wavelet transformation (DWT) to systematically extract key time-based features related to the exchange rate behavior over different time horizons. To evaluate the performance of the proposed model the authors used a dataset containing instances about currency crises between 1992 and 2015. Experimental results showed that the proposed RF-DWT model achieved a high prediction accuracy between 89\%-90\%, outperforming the LR model which achieved an accuracy between 84\%-85\%.\\
\indent Similarly, \citet{s81_c} also proposed the combination of RF and DWT for bi-annual currency crises prediction based on the exchange market pressure (EMP) index. To that end, the authors used a dataset covering 101 industrial and developing countries between 1994-2018 from the International Financial Statistics of the International Monetary Fund (IMF). Experimental results showed that the proposed model achieved a prediction accuracy of close to 73\%, outperforming other models by at least 5\%. This highlighted the potential of the proposed bi-annual forecasting model in providing guidance for both policy makers and investors to detect currency risks.\\
\indent In contrast, \citet{s81_d} proposed a deep learning model to predict currency crises. More specifically, the authors proposed the use of deep neural decision trees (DNDT) and compared its performance to other widely adopted methodologies. To compare the performance of the different models, the authors used a dataset consisting of 162 developed, emerging, and developing countries with information between 1970–2017. Experimental results showed the proposed model achieved a training prediction accuracy ranging between 98\%-99\% and a testing accuracy between 97\%-99\% across multiple geographical regions. This highlights the potential of deep learning models for accurate currency crises prediction.\\ 
\subsubsection{Research Opportunities:}
\indent Similar to other opportunities in the banking and finance sector, there are multiple research opportunities in which ML can play a role for financial crises prediction. One opportunity is investigating the performance of existing models in predicting financial crises in specific fields rather than just at the macro/country level. For example, study the effectiveness of existing models in predicting crises in the housing field since such models can be extremely helpful for real-estate developers and landlords. Another potential opportunity is exploring the effectiveness of different classification models in predicting such crises and studying their complexity. A third potential research opportunity is exploring other deep learning models such as CNN and RNN to investigate their performance given the promising results achieved by other deep learning architectures. \\
\indent Table \ref{ML_finance_table} briefly summarizes the challenges, previous works, and potential research opportunities of ML within the banking and finance sector.
\begin{table}[!tbp]
	\centering
	\caption{\label{ML_finance_table} Challenges, Previous Works, and Research Opportunities within Banking and Finance Field}
	\scalebox{0.9}{
	\begin{tabular}{|p{1.5cm}|p{5.5cm}|p{4.5cm}|}
		\hline
		Challenge & Previous Work & Research Opportunity  \\ \hline
		\multirow{3}{1.5cm}{Credit Risk Assessment}&Proposed the combined use of unsupervised and supervised ML models to assess the credit risk of individuals. \citep{s78_a}& - Explore hyper-parameter optimization of the ML models considered to improve their performance. \\ \cline{2-2}
		&Proposed a hybrid ensemble ML model composed of two ensemble learning models and seven classification models to assess the credit risk of SMEs in supply chain finance \citep{s78_b}&- Explore the performance of different models that can make short, medium, and long term risk prediction. \\ \cline{2-2}
		&Proposed DBN composed of RBM and Softmax classifier to predict the credit risk of SMEs working in the online supply chain space&- Consider other deep learning models and architectures such as CNN and RNN to investigate the improvement in the credit assessment accuracy. \\ \hline
		\multirow{4}{1.5cm}{Bankruptcy Prediction}&Compared performance of optimized SVM with ANN and other ML techniques to predict institution bankruptcy \citep{s80} & - Study the impact of other ML models and kernels in performing the bankruptcy prediction. \\ \cline{2-2}
		&Explored four different ML models including SVM, DT bagging, DT boosting, and RF models for bankruptcy prediction \citep{s80_a}&- Investigate the performance of different optimization models and meta-heuristics to study the potential trade-off between performance improvement and computational complexity. \\ \cline{2-2}
		&Proposed the use of ensemble learning models in combination with feature selection as part of their bankruptcy prediction models \citep{s80_b}& - Explore other deep learning models and architectures such as RNN to investigate their performance in comparison with CNN models proposed in the literature.\\ \cline{2-2}
		&Proposed the use of CNN model to predict bankruptcy based on textual data in conjunction with accounting-based ratio and market-based variables \citep{s80_c}&\\ \hline
		\multirow{4}{1.5cm}{Currency Crises Prediction}& Proposed the use of SVM model to predict currency crises \citep{s81_a}&- Investigate the performance of existing models in predicting financial crises in specific fields rather than just at the macro/country level.\\ \cline{2-2}
		&Proposed the combination of RF and wavelet transformation to predict currency crises \citep{s81_b}&- Explore the effectiveness of different classification models in predicting such crises and studying their complexity.\\ \cline{2-2}
		&Proposed a combined RF-DWT model to predict currency crises \citep{s81_c}& - Explore other deep learning models such as CNN and RNN to investigate their performance. \\ \cline{2-2}
		&Proposed the use of deep neural decision trees (DNDT) and compared its performance to other widely adopted methodologies \citep{s81_d}&\\ \hline
	\end{tabular}}
\end{table}
\section{Social Media}\label{Social_Media}
\indent Another emerging area in which ML has been playing a major role is the area of social media. \citet{s82} defines social media as ``the means of interactions among people in which they create, share, and/or exchange information and ideas in virtual communities and networks''. The first form of social media appeared in 1979 when USENET created a decentralized system of discussion boards \citep{s83}. Since then, the Internet has advanced well beyond discussion boards where interactions occur in text only. In the early 21st century, many websites were launched that provide users with a platform to not only communicate via text, but also to share videos and/or photos \citep{s12}.\\
\indent According to \citet{s82}, eight of the most popular social media platforms are Facebook, Twitter, Youtube, Vimeo, Flickr, Instagram, Snapchat, and LinkedIn. As of 2019, there are 2.77 billion social media users worldwide, with it being projected that in 2021 there will be 3.02 billion social media users worldwide \citep{s84}. There are 2.5 quintillion bytes of social media data created each day \citep{s85}. Furthermore, every minute of the day Snapchat users share 527,760 photos, 456,000 tweets are sent on Twitter, Instagram users post 46,740 photos, and Facebook users post 510,000 comments and 293,000 status updates \citep{s85}. Also, on Facebook more than 300 million photos are uploaded per day \citep{s85}. In conclusion, the vast amount of data produced by social media cannot be processed by humans. Hence, social media provides another area of opportunity for the use of ML. In this section, how ML techniques can be applied to social media data in order to make discoveries in the fields of pharmacovigilance, vaccine sentiment analysis, and politics will be discussed.
\subsection{Pharmacovigilance}
\subsubsection{Challenge Description:}
\indent One way that social media data is being used is in pharmacovigilance. Pharmacovigilance (PhV) is defined as ``the science and activities relating to the detection, assessment, understanding, and prevention of adverse effects or any other drug-related problem'' \citep{s86}. Adverse effects, also known as Adverse Drug Reactions (ADRs) are harmful reactions that are caused by the intake of medication \citep{s87}. ADRs have led to millions of deaths and hospitalizations and cost nearly seventy-five billion dollars annually. Governmental agencies such as the U.S. Food and Drug Administration (FDA) and the European Medicines Agency (EMA), along with international organizations such as the World Health Organization (WHO) engage in pharmacovigilance by requiring manufacturers to report adverse events \citep{s88}. These agencies also encourage voluntary reporting by healthcare professionals and the public. However, there is no guarantee that healthcare professionals or the public will report ADRs. Furthermore, when ADRs are voluntarily reported, the information may not be timely, may be incomplete, duplicated, under-reported or over-reported. Due to the limited quantity and lack of quality of voluntarily reported ADRs, it has become necessary to supplement voluntary reports with other data forms. For example, information about ADRs can be acquired from health-related social networks such as DailyStrength or on social media sites such as Twitter and Facebook. \\
\indent Although these sites provide a vast amount of data for potential ADR detection, it is impossible for a human to analyze all of the data. Hence, natural language processing (NLP) and ML algorithms have been used to process the data \citep{s87,s88}. A survey of the literature shows that NLP techniques are commonly used to analyze social media data for ADRs via text classification using lexicon-based approaches. Furthermore, SVM, NB, and Maximum Entropy algorithms have been used to classify text. While these approaches provide a novel opportunity for collecting data about ADRs, there are still many challenges to using these approaches. For example, pure lexicon-based approaches are often impeded by consumers not using technical terms, misspelling words, using abbreviations, or sentence structure irregularities. Furthermore, when supervised learning approaches are used, they require substantial amounts of data to be manually annotated, often by a domain expert. That being said, researchers have begun to use partially supervised (semi-supervised) algorithms in order to reduce the amount of annotated data that is required \citep{s103}. 
\subsubsection{Previous Works:} 
\indent \citet{s91} utilized ML on tweets in order to discover mentions of ADRs. However, in their work the authors found that false positive errors were occurring due to non-ADR extracted terms being classified as ADRs. As an example, the authors discuss the username TScpCancer, which was classified as an ADR even though the word cancer is being used as a name in this context.\\ 
\indent \citet{s98} used ML techniques on social media data to automatically classify drugs into either a normal category or a blackbox category (blackbox is a category of drugs that the FDA has identified as having serious or life-threatening safety concerns). The authors’ approach showed promise at classifying social media comments as ADRs or non-ADRs. However, their approach was only marginally successful at classifying drugs into the normal or blackbox categories. The authors believe that they encountered this challenge due to their limited annotated dataset. Furthermore, the authors found it challenging to distinguish true signals from the noisy social media text data. \\
\indent \citet{s99} proposed a SVM-based framework for integrated and high-performance patient reported adverse drug event extraction from social media. More specifically, the authors used data collected from four major diabetes and heart disease forums in the United States and applied various natural language processing models to create the lexicon-based datasets to be fed to the SVM classifier. Experimental results showed that the proposed model achieved a high precision ranging between 91\%-94\% for ADR and between 79\%-87\% for medical events.\\
\indent Similarly, \citet{s100} proposed the use of SVM to accurately identify ADR posted by patients on social media platforms. To that end, the authors considered two datasets, namely the CSIRO Adverse Drug Event Corpus (CADEC) and the Twitter Corpus. From these datasets, a set of context-level and entity-level features were extracted and provided as an input to the proposed linear SVM model. Experimental results showed that the proposed model achieved a high precision ranging between 81\%-84\% for the CADEC corpus dataset and between 64\%-73\% for the Twitter corpus dataset.\\
\indent In contrast, \citet{s101} proposed a scalable deep-learning model to analyze and identify ADRs in social media posts. More specifically, the authors proposed the use of RNN that labels words in an input sequence with ADR membership tags. To that end, the authors used a Twitter corpus dataset to evaluate the performance of the proposed RNN-based framework. Experimental results showed that the authors' model outperformed other traditional models such as the baseline lexicon matching (LM) system and conditional random field model (CRF) by achieving an F1-measure of 0.755 for ADR identification compared to 0.63 and 0.65 for the LM and CRF models respectively.
\subsubsection{Research Opportunities:}
\indent Although many previous work has utilized ML for analyzing social media posts concerning drugs and medications, there still exist many further opportunities. One potential opportunity is examining the effectiveness of ML classification in modeling the contextual and semantic features of tweets. Another opportunity worth exploring is enriching the ADR lexicon datasets so that the sentiment analysis of tweets and social media posts becomes more accurate. Another potential research opportunity is performing temporal analyses to mine drug-ADR patterns and investigate ADRs related to the interaction of drugs taken by patients. Also, researchers can explore more complex classification models such as hidden markov models (HMM) to distinguish between symptoms and side-effects mentioned in the posts. Moreover, a transfer learning model can be explored to transfer knowledge from one classification domain to another, \textit{i.e.} potentially from one drug to another or from one platform to another.
\subsection{Social Media and Vaccines}
\subsubsection{Challenge Description:}
\indent Another way that ML can be used to gather information from social media data is by determining people beliefs, thoughts, and feelings about various vaccines. In recent years it has been observed that some individuals and/or groups have negative opinions about the safety and value of vaccines, and these negative opinions are being expressed online via social media. These negative opinions may influence some people’s decisions to receive vaccines or to vaccinate their children \citep{s106,s107,s108,s109,s110}. In the past decade, in the United States and other countries, there has been an increase of parents refusing to vaccinate their children due to their concerns about the safety of vaccines. Vaccine refusal for one’s self or one’s child can result in unnecessary harm or even death. One way that scientists and researchers are combating the anti-vaccination movement is by analyzing social media data with ML algorithms in order to understand how negative opinions about vaccines spread through social media. Once these patterns are understood, scientists and researchers hope that they can combat the spread of misinformation.
\subsubsection{Previous Works:} 
\indent \citet{s106} hypothesized that when Twitter users were exposed to negative opinions about human papillomavirus (HPV) vaccines in Twitter communities that these users would subsequently express the negative opinions that they were exposed to by re-posting similar negative opinions. In order to examine their hypothesis, the authors analyzed temporal sequences of messages posted on Twitter (tweets) related to HPV vaccines and the social connections between users. The researchers’ dataset was collected between October 2013 and April 2014. The dataset consisted of 83,551 tweets written in English that included terms related to HPV vaccines. Furthermore, the social connections (N = 957,865) of the 30,621 users who posted or reposted the tweets were examined to see if they also posted or reposted such tweets. In order to analyze this large dataset, the authors utilized a supervised ML approach to classify the tweets. This approach required the researchers to first manually label a random sample of tweets. Then, the labeled tweets were used to train a ML classifier to recognize similar patterns in the remaining tweets. More specifically, the classifier was an ensemble of four classifiers that used the content of the tweets (the words and word combinations in the tweets themselves) or the social relations between users (the users followed by the user responsible for the tweet) in order to classify the sentiment of the tweets. The sentiment of the users’ tweets about HPV vaccines was classified either as negative or neutral/positive. When the four classifiers were trained and tested in a 10-fold cross validation, their accuracy ranged between 87.6\% and 94.0\%. The researchers concluded that Twitter users who were more often exposed to negative opinions about HPV vaccines were more likely to subsequently post negative tweets about HPV vaccines.\\ 
\indent Similarly, \citet{s110} proposed a hierarchical SVM-based model to predict tweet sentiments about HPV vaccines. To that end, the authors collected tweets written in English containing HPV vaccines-related keywords in the time between November 2, 2015 and March 28, 2016. Experimental results showed that the proposed model achieved a high precision ranging between 71\%-78\%. Moreover, the model also showed particularly high precision in identifying negative sentiments pertaining to HPV vaccine safety with a value of around 80\%.\\
\indent \citet{s109} used natural language classifiers to examine and analyze data from Twitter in order to track flu vaccinations over time, as well as by geography and gender. The researchers collected a dataset of 1,007,582 tweets. From this dataset, the researchers created a training dataset by annotating a random sample of 10,000 tweets. After testing various classifiers, the researchers chose the best-performing classifier, namely LR, and used it in the rest of their experiments. When the researchers compared the results of their algorithm to a published government survey data about vaccination from the US Centers for Disease Control and Prevention (CDC), they found that their results were highly correlated with the CDC’s data (r = 0.90). These results suggest that ML algorithms can be applied to Twitter data in order to track people’s attitudes and behaviors about flu vaccinations.
\subsubsection{Research Opportunities:}
\indent As evident by the different research works discussed above, ML has great potential as it can be used to examine large amounts of social media data in order to track and determine how social media users may influence each other’s opinions of vaccines. While these studies have shown great potential for the use of ML, there are still some limitations that offer possibilities for future research. One limitation that could use further development is that social media users’ connections change over time and this may not be reflected in data that is taken from a set time period. Therefore, it is important to develop adaptive ML models that can change with the social connection changes of the social media platform. Another potential opportunity is creating new datasets with updated vocabulary to better track the sentiment of users based on the non-standard abbreviations, slang and phrases commonly used on social media platforms. A third research opportunity worth exploring is to study the performance of deep learning architectures such as CNNs and RNNs for sentiment analysis of various vaccines. This is particularly important given the promising performance illustrated by such architectures in analyzing social media posts.
\subsection{Social Media and Politics}
\subsubsection{Challenge Description:}
\indent Moving beyond health-related topics, ML techniques can also be applied to social media in order to collect, monitor, analyze, summarize, and visualize politically relevant information \citep{skb_sm1}. In recent years, social media platforms such as Twitter and Facebook have been used to increase political participation. For example, social media users publicly spread information about their political opinions on Twitter and political institutions have begun to use Facebook pages or groups to engage with citizens \citep{s111}. Furthermore, politicians and political parties are interested in social media data, because they can benefit from understanding what the public thinks about them \citep{s112}. Due to their interest in public opinion about politics, politicians or political parties may monitor social media data in order to detect social media content that is directly or indirectly associated with them. Furthermore, the monitoring of political social media data is also important because it may provide information about potential political crises or scandals. Additionally, the spread of political information through social networks can lead to administrative, political and societal changes. For example, social media played a central role in shaping political debates in the Arab Spring (a series of pro-democracy protests, uprisings, and armed rebellions that spread across North Africa and the Middle East beginning in the spring of 2011 \citep{s114}).\\
\indent A common method that is used for the detection and analysis of political social media content is opinion mining (also known as sentiment analysis). The process of political opinion mining consists of collecting text that contains political opinions (or sentiments) and extracting attributes and components about a specific political feature from said text, then determine whether the text is positive, negative or neutral.  
\subsubsection{Previous Works:} 
\indent \citet{s117} have implemented sentiment analysis on tweets. They were interested in the predictive power of social media. In their study, the authors analyzed 32 million tweets related to the 2012 US presidential election using a combination of ML techniques. The authors implemented a Twitter crawler from September 29, 2012 until November 16, 2012 using keywords such as Barack Obama, Mitt Romney, US election, Paul Ryan, and Joe Biden. Their results were numerous. Firstly, the authors' results (that Obama was leading in Twitter for the 2012 US presidential election) matched with the outcome of the election. Secondly, the authors found that by analyzing geo-tweets (tweets with a geo-tag) with geographical sentiment analysis, they were able to uncover the popularity of candidates across the US states. Thirdly, the authors work demonstrated that LDA is a powerful unsupervised algorithm when combined with the NB classifier as it was able to ``predict'' the result of the 2012 US election. Hence, the authors have presented a system of mining social media data that may be used for predicting future events.\\
\indent In a similar fashion, \citet{s118} proposed the use of ML models to predict the results of the 2016 US presidential election based on sentiment analysis of the corresponding tweets. To that end, the authors proposed the use of NB and SVM models to classify the tweets about Donald Trump and Hilary Clinton. The performance of the model was evaluated using a dataset collected through Twitter between March 16th-17th, 2016 which was later labeled manually. Experimental results showed that the proposed models achieved a sentiment prediction accuracy ranging between 97\%-99\% with an F1-score between 0.94-0.97, highlighting the potential of ML models in predicting election results based on the sentiment analysis of tweets.\\
\indent \citet{s119} also proposed the use of ML models to predict the results of the 2019 Nigerian presidential election by comparing three lexicon-based classifiers (VADER, VADER-EXT, and Textblob) and five ML-based classifiers (SVM, LR, NB, stochastic gradient descent SGD, and RF). To evaluate the performance of the different models, the authors collected 118,421 posts between January 1 and February 22, 2019. Experimental results showed that the VADER-Ext approach outperformed the other two lexicon-based approaches by achieving a precision of approximately 81\%. In a similar fashion, it was shown that the LR method achieved the highest accuracy and precision of 77\% and 78\% respectively among the different ML models. \\
\indent On the other hand, \citet{s120} extended the concept of sentiment analysis by using deep learning models to predict multiple local election results rather than the national election. To that end, the authors proposed the use of recursive neural tensor network (RNTN) to analyze the sentiment shown in various Twitter posts about the 2018 US Midterm elections. The performance of the proposed model was evaluated using a manually collected dataset consisting of approximately 800 tweets. Experimental results showed that the proposed model achieved a high prediction accuracy as it predicted an advantage of 9.2\% for the Democratic candidate compared to the actual advantage which was measured to be 8.6\%. As such, it was shown that the proposed model indeed has great potential in accurately predicting multiple local election results.
\begin{figure}[!t]
	\centering
	\includegraphics[scale=0.4]{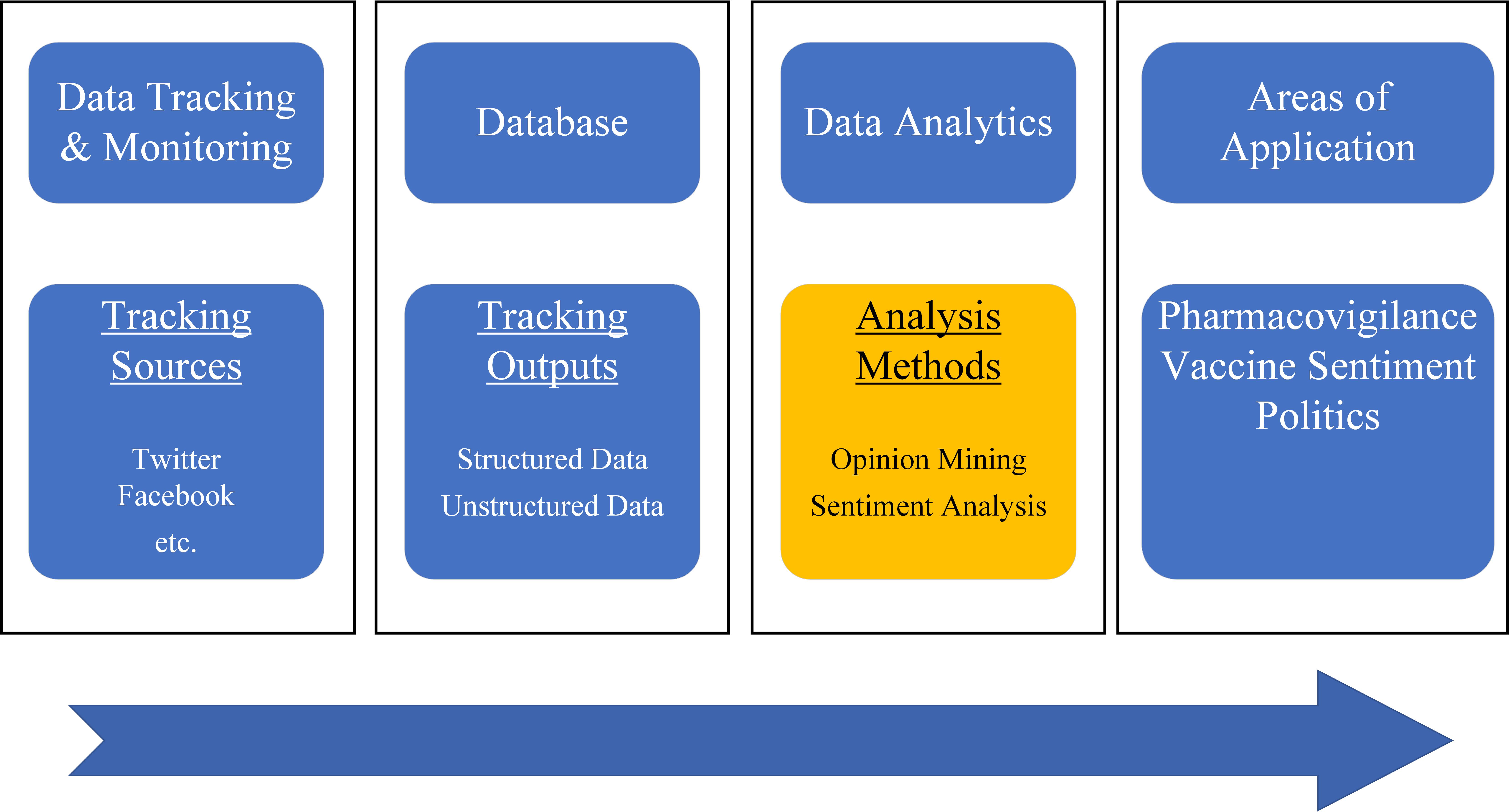}
	\caption{Potential ML-based Social Media Analytics Framework}
	\label{ML_in_socialmedia} 
\end{figure}
\subsubsection{Research Opportunities:}
\indent Again, there are still many research opportunities in which ML can play a role as part of a politics sentiment analysis frameworks. One such opportunity is investigating other ML algorithms such as SVM and ANN given their previous success in determining linguistic features for opinion classification. Another potential opportunity is collecting more features such as swear words, sarcasm, and negative and conditional detection as well as contextual clues features to make the sentiment analysis framework more accurate and effective. A third opportunity is to consider other deep learning models such as CNN and RNN to compare their performance with the currently proposed deep learning models. This is crucial given the promising results achieved by deep learning models such as RNTN model.\\ 
\indent Table \ref{ML_SM_table1} summarizes some of the different challenges and research opportunities of ML within the social media field. Moreover, Figure \ref{ML_in_socialmedia} provides a visualization of how these topics fit into a Social Media Analytics Framework.
\begin{table}[h]
	\centering
	\caption{\label{ML_SM_table1} Challenges, Previous Works, and Research Opportunities within Social Media Field}
	\scalebox{0.9}{
	\begin{tabular}{|p{2.5cm}|p{4.5cm}|p{4.5cm}|}
		\hline
		Challenge & Previous Work & Research Opportunity \\ \hline
		\multirow{5}{2.5cm}{Pharmacovigilance}&Utilized ML on tweets in order to discover mentions of ADRs \citep{s91}& - Examine the effectiveness of ML classification in modeling the contextual and semantic features of tweets.\\ \cline{2-2}
		&Used ML techniques on social media data to automatically classify drugs into either a normal category or a blackbox category \citep{s98}& - Enrich the ADR lexicon datasets so that the sentiment analysis of tweets and social media posts become more accurate.\\ \cline{2-2}
		&Proposed a SVM-based framework for integrated and high-performance patient reported adverse drug event extraction from social media \citep{s99}& - Perform temporal analyses to mine drug-ADR patterns and investigate ADRs related to the interaction of drugs taken by patients.\\ \cline{2-2}
		& Proposed the use of SVM to accurately identify ADR posted by patients on social media platforms \citep{s100}& - Explore more complex classification models to distinguish between symptoms and side-effects mentioned in the posts.\\ \cline{2-2}
		&Proposed a scalable RNN model to analyze and identify ADRs in social media posts \citep{s101} & - Explore transfer learning models to transfer knowledge from one classification domain to another.\\ \hline
		\multirow{3}{2.5cm}{Social Media and Vaccines}&Utilized a supervised ML approach to classify vaccine-related tweets \citep{s106}& - Develop adaptive ML models that can change with the social connection changes of the social media platform.\\ \cline{2-2}
		&Used SVM model on Twitter data in order to assess HPV vaccination sentiments \citep{s110}& - Create new datasets with updated vocabulary to better track the sentiment of users based on the non-standard abbreviations, slang and phrases commonly used on social media platforms.\\ \cline{2-2}
		&Used natural language classifiers to examine and analyze data from Twitter in order to track flu vaccinations over time, geography, and gender \citep{s109}& - Study and compare the performance of deep learning architectures such as CNNs and RNNs. \\ \hline
		\multirow{4}{2.5cm}{Social Media and Politics}&Created a ML-NLP engine that implemented a NB classifier for sentiment analysis, and the Latent Dirichlet Allocation (LDA) algorithm for topic modeling \citep{s117}& - Investigate other ML algorithms such as SVM and ANN given their previous success in determining linguistic features for opinion classification. \\ \cline{2-2}
		&Proposed the use of NB and SVM models to predict the results of the 2016 US presidential election based on corresponding tweets \citep{s118}& - Collect more features such as swear words, sarcasm, and negative and conditional detection as well as contextual clues features to make the sentiment analysis framework more accurate and effective.\\ \cline{2-2} 
		&Compared the performance of three lexicon-based and five ML-based models in predicting the results of the 2019 Nigerian presidential election \citep{s119}& - Consider other deep learning models such as CNN and RNN to compare their performance with the currently proposed deep learning models.\\ \cline{2-2}
		&Proposed the use of RNTN deep learning model to predict the results of the 2018 US Midterm elections \citep{s120}&\\ \hline
	\end{tabular}}
\end{table}
\begin{figure}[t]
	\centering
	\includegraphics[scale=0.35]{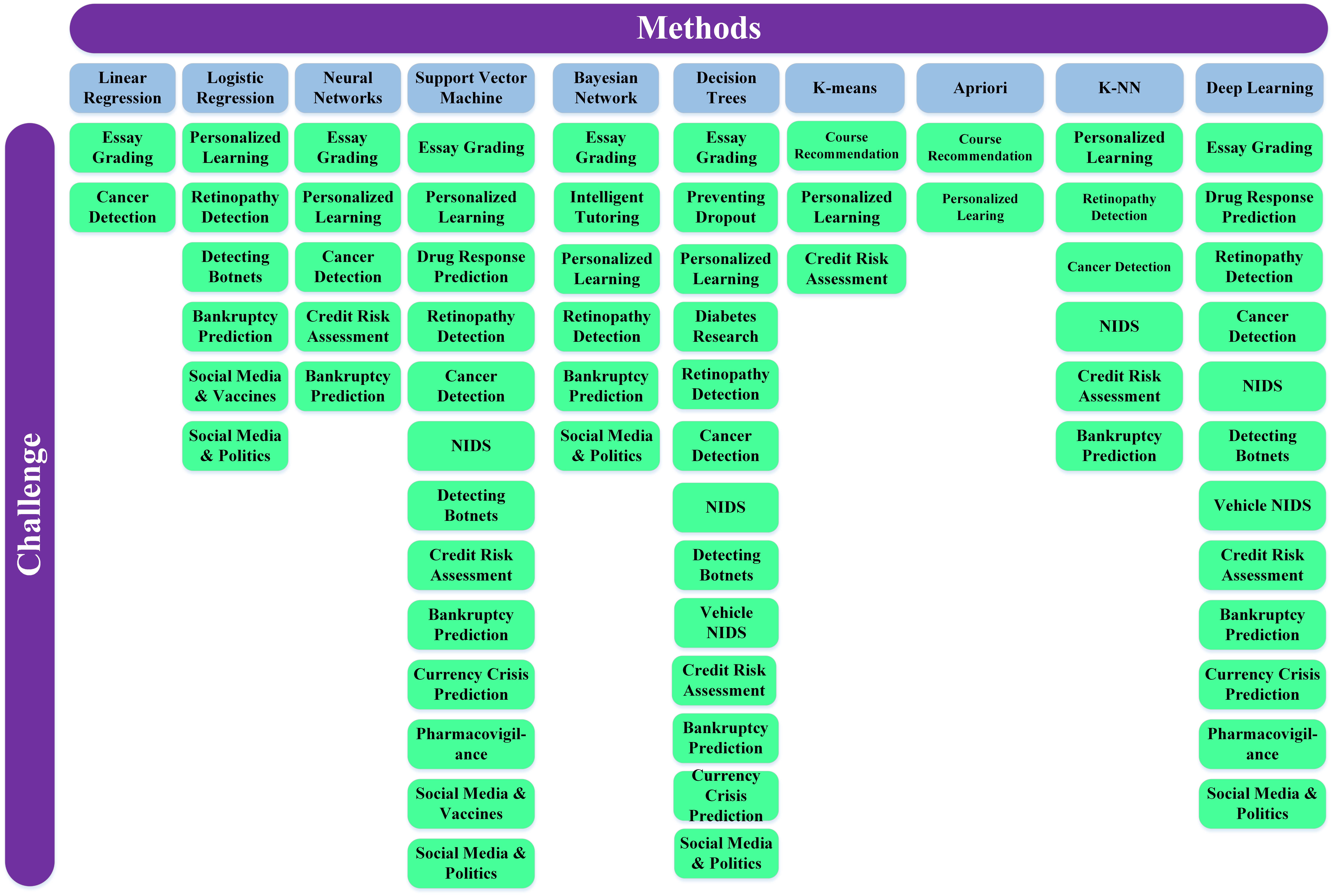}
	\caption{Summary of Challenges and ML Techniques}
	\label{ML_app_technique_summary} 
\end{figure}
\section{Conclusion}\label{conclusion}
\indent The availability and popularity of the Internet and related technologies has resulted in large amounts of data being available for analyses. However, humans do not possess the cognitive capabilities to understand such large amounts of data. Machine learning (ML) provides a way for humans to process large amounts of data and come to conclusions about the data. ML has applications in various fields. This review focused on some of the fields and applications such as education, healthcare, network security, banking and finance, and social media. These fields each have their own unique challenges. However, ML can provide solutions to these challenges, as well as create further research opportunities. Accordingly, this work briefly described some of the challenges facing the aforementioned fields and surveyed some of the previous literature works that focused on them. Moreover, it presented several research opportunities on the role and potential of using ML to address these challenges. Figure \ref{ML_app_technique_summary} summarizes the challenges and previous/potential ML techniques that addressed/can address them respectively.\\

\small
\textbf{Acknowledgments} This study was funded by Ontario Graduate Scholarship (OGS) Program.\\
\textbf{Conflict of Interest} The authors declare that they have no conflict of interest.\\
\textbf{Informed Consent} This study does not involve any experiments on animals.\\

%
%

\bibliographystyle{spbasic}      
\bibliography{ch3-ref}   


\end{document}